\documentclass[10pt, a4paper]{article}

\usepackage{subfigure}
\usepackage{subcaption}
\usepackage{algpseudocode}
\usepackage{amsmath, amssymb, amsthm, bm}
\usepackage{booktabs, multirow}
\usepackage{algorithm}
\DeclareMathOperator*{\argmax}{arg\,max}
\DeclareMathOperator*{\argmin}{arg\,min}

\usepackage[final]{lrec2026} % this is the new style
\title{Detecting and Rectifying Noisy Labels: \\ A Similarity-Based Approach}

\name{Dang Huu-Tien$^{\dagger}$, Minh-Phuong Nguyen$^{\dagger}$, and Naoya Inoue$^{\dagger,\ddagger}$}

\address{$^{\dagger}$JAIST,\; $^{\ddagger}$RIKEN \\
         \{tiendh, phuongnm, naoya-i\}@jaist.ac.jp\\}
\abstract{
Label noise in datasets could significantly damage the performance and robustness of deep neural networks (DNNs) trained on these datasets. As the size of modern DNNs grows, there is a growing demand for automated tools for detecting such errors. In this paper, we propose \textit{post-hoc, model-agnostic} noise detection and rectification methods utilizing the penultimate feature from a DNN. Our idea is based on the observation that \textit{the similarity between the penultimate feature of a mislabeled data point and its true class data points is higher than that for data points from other classes}, making the probability of label occurrence within a tight, similar cluster informative for detecting and rectifying errors.
Through theoretical and empirical analyses, we demonstrate that our approach achieves high detection performance across diverse, realistic noise scenarios and can automatically rectify these errors to improve dataset quality. 
Our implementation is available at \url{https://anonymous.4open.science/r/noise-detection-and-rectification-AD8E}.
 \\ \newline \Keywords{Noisy label detection and rectification, post-hoc, model-agnostic, DNN, LLM} }

\begin{document}

\maketitleabstract

\section{Introduction}
While the majority of knowledge in Artificial Intelligence (AI) systems is learned through unsupervised learning, supervised learning is an indispensable step in building strong AI systems (c.f. the LeCun's cake). For LLMs such as GPTs~\cite{gpt3}, Llama~\cite{llama, llama2}, and Gemini~\cite{gemini}, supervised learning accounts for only a small fraction of the total computation budget but has a significant impact on the models' performance. 

The efficacy and resilience of supervised models are susceptible to degradation due to noise in datasets. Noise (mislabeled) samples could result from mistakes in the labeling process or human disagreement~\cite{goh2023activelab, goh2022utilizing}. Human error rates on real-world datasets could range from a few to tens of percent~\cite{beyer2020we, northcutt2021pervasive}. 
% For example, error rate on the ImageNet~\cite{deng2009imagenet} or CIFAR-100~\cite{cifar-100} is over $5.0\%$~\cite{beyer2020we, northcutt2021pervasive}. 
For more challenging datasets where expert knowledge is required, the error rate could be even higher.
Recent research~\cite[\textit{e.g.,}][]{lima, textbooksareallyouneed} finds that high-quality training data significantly improves performance while reducing the training cost by orders of magnitude. 
The large scale of modern datasets makes manual data inspection prohibitively expensive. The immense complexity of deep models makes it extremely hard to reliably and efficiently predict the effect a data point has on the performance of a deep model \cite{basu2021influence}. 

The need for automated tools for improving the quality of supervised learning data is rising as datasets and LLMs are getting larger at an unprecedented speed.
Previous works \cite{VanAnh, Thang} showed that removing noise from the training set improves the performance of AI models trained on that dataset. Automatic label noise rectification, however, is an underexplored topic. In this paper, we make the following contributions:
\begin{enumerate}
    \item[(1)] We first provide a theoretical analysis from a similarity perspective, showing why the similarity between the penultimate feature of a mislabeled data point and its true class data points is often larger than that for data points from other classes (Section~\ref{theoretical_analysis}).
    \item[(2)] Inspired by this observation, we develop a simple yet effective similarity-based approach with two representative methods for detecting and rectifying label noise in large-scale datasets (Section~\ref{sec:method}). Our methods are \textit{posthoc} and \textit{model-agnostic}, that is, they can be applied to any DNN architectures.
    \item[(3)] Extensive experiments demonstrate the superior performance of the proposed methods over gradient-based and confidence-based approaches, across diverse noise scenarios and scales~(Section~\ref{sec:result}). 
    %Furthermore, our approach generalizes to zero-shot scenarios with modern LLMs.
\end{enumerate}

\section{Background and Related Work}
\label{sec:append-how-prod}
\paragraph{Notation.} Let us define $\mathbf{z} = (\mathbf{x}, \mathbf{y}$) a data point, where $\mathbf{x} \in \mathcal{X}$ is an input and $\mathbf{y}\in \mathcal{Y}$ is a target output. Let $\mathcal{D} = \{\mathbf{z}^{(i) }=(\mathbf{x}^{(i)},\mathbf{y}^{(i)})\}_{i=1}^{n}$ be a $N$-class (potential noisy) training dataset of $n$ data points. 
Let $f_{\bm \theta}: \mathcal{X} \mapsto \mathcal{Y}$ be a model parameterized by $\bm\theta$; $\hat{\bm\theta} = \argmin_{\bm\theta}\frac{1}{n}\sum_{i=1}^n\ell(\mathbf{z}^{(i)}, \bm\theta)$ are optimal parameters of $f_{\bm\theta}$ measured on $\mathcal{D}$, where $\ell: \mathcal{Y}\times \mathcal{Y} \mapsto \mathbb{R}_{+}$ be the loss function. In this paper, $\bm g_{\hat{\bm\theta}}(\mathbf{z}^{(i)}) = \nabla_{\theta}\ell(\mathbf{z}^{(i)}, \hat\theta)$ is denoted as the gradient of the loss at $\mathbf{z}^{(i)}$ with respect to (w.r.t) $\bm\theta$ in model $f_{\hat{\bm\theta}}$. 
%$||\cdot||$ denote the Euclidean norm, $\langle \cdot,\cdot \rangle$ denote the inner product.
%A noise detection model $g: \mathcal{X} \times \mathcal{Y} \rightarrow \mathbb{R}$ assigns a score to each data point in $\mathcal{D}$, indicating its likelihood of being noisy.
\subsection{Confidence-Based Approach} 
Confidence-based methods are based on the notion of \textit{confident learning}~\cite{CL} that derives label quality measurements by using predicted probability distribution~\cite{wang2022detecting, selfconfidence, thyagarajan2022identifying}. Low confidence serves as a heuristic indicating the likelihood of label noise. Given a data point $\mathbf{z}$ with output label $\mathbf{y} = (y_1,..., y_k, ...y_N)$, the model's predicted probabilities is $\mathbf{p} = (p_{1},...,p_k,..., p_N)$ over $N$ classes. \citet{CL} proposed three label quality scoring methods: 

\noindent \textbf{Self-Confidence} (SC) refers to the estimated probability that the input $\mathbf{x}$ belongs to the class associated with its given label $k$: 
\begin{align}
    \textnormal{SC}(\mathbf{z}, \mathbf{p}) = p_{y_k}; \;k \in \{1, 2, ..., N\}
\end{align}
% $\textnormal{SC}(\mathbf{z}, \mathbf{p}) = p_{y_k}$, for $k \in \{1, 2, ..., N\}$. 
\textbf{Normalized-Margin} (NM) is the quantified difference between the model's estimated probability of the given label and the probability of the most likely class: 
\begin{align}
    \textnormal{NM}(\mathbf{z}, \mathbf{p}) = p_{y_k} - p_{y_j{^*}}; \; j^{*} = \argmax_{j\neq k \in \{1, 2, ..., N\}} p_{y_j}
\end{align}
\textbf{Confidence-Weighted Entropy} (CE) is the ratio of SC score and the normalized entropy: 
\begin{align}
    \textnormal{CE}(\mathbf{z}, \mathbf{p}) = \frac{p_{y_k}}{S_{N}(\mathbf{p})};  S_N(\mathbf{p}) = -\frac{1}{\textnormal{log}{N}} \displaystyle\sum_{n=1}^{N}p_n \log(p_n)
\end{align}
\subsection{Gradient-Based Approach}
% \noindent \textbf{Gradient-based approach.} 
\citet{IF} use \textbf{Influence Function} (IF)---a concept from robust statistics~\cite {if_robuststats}---for measuring the influence of a training data point on others in a DNN without retraining. Several IF variants with faster measures have been proposed. \citet{VanAnh} proposed a way to adapt IF and its variants, that is, \textbf{Gradient Dot Product} (GD;~\citet{gd_gc}), \textbf{Gradient Cosine} (GC;~\citet{gd_gc}), and \textbf{Tracing Gradient Decent} (TracIn;~\citet{tracin}), for identifying errors in large-scale source code datasets. The idea is that the gradients of noisy data points exhibit significantly large magnitudes and are opposite in direction to the gradients of normal data points. The algorithm computes the influence score of each data point in the noisy dataset with data points in a reference set. A more negative influence score means it is more likely to be a noisy data point. \citet{Thang} use class information to improve the performance and stability of these gradient methods. We summarize the gradient-based approach as follows:
\begin{align}
    \textnormal{IF}(\mathbf{z}^{(i)}, \mathbf{z}^{(j)}) &= -\frac{1}{n}\bm{g}_{\hat\theta} (\mathbf{z}^{(i)})^{\top} \bm{H}_{\hat{\bm\theta}}^{-1} \bm{g}_{\hat{\bm\theta}}(\mathbf{z}^{(j)}) \\
    \textnormal{GD}(\mathbf{z}^{(i)}, \mathbf{z}^{(j)}) &= \left\langle \bm{g}_{\hat{\bm\theta}}(\mathbf{z}^{(i)}), \bm{g}_{\hat{\bm\theta}}(\mathbf{z}^{(j)}) \right\rangle \\
    \textnormal{GC}(\mathbf{z}^{(i)}, \mathbf{z}^{(j)}) &= \frac{\left\langle \bm{g}_{\hat{\bm\theta}}(\mathbf{z}^{(i)}), \bm{g}_{\hat{\bm\theta}}(\mathbf{z}^{(j)}) \right\rangle}{||\bm{g}_{\hat{\bm\theta}}(\mathbf{z}^{(i)})||||\bm{g}_{\hat{\bm\theta}}(\mathbf{z}^{(j)})||}\\
    \textnormal{TracIn}(\mathbf{z}^{(i)}, \mathbf{z}^{(j)}) &= \sum_{t=1}^T \eta_t \left\langle \bm{g}_{\bm\theta^{(t)}}(\mathbf{z}^{(i)}), \bm{g}_{{\bm\theta^{(t)}}}(\mathbf{z}^{(j)}) \right\rangle
\end{align}
% (1) $\textnormal{IF}(\mathbf{z}^{(i)}, \mathbf{z}^{(j)}) = -\frac{1}{n}\bm{g}_{\hat\theta} (\mathbf{z}^{(i)})^{\top} \bm{H}_{\hat{\bm\theta}}^{-1} \bm{g}_{\hat{\bm\theta}}(\mathbf{z}^{(j)})$, 
% (2) $\textnormal{GD}(\mathbf{z}^{(i)}, \mathbf{z}^{(j)}) = \left\langle \bm{g}_{\hat\theta}(\mathbf{z}^{(i)}), \bm{g}_{\hat{\bm\theta}}(\mathbf{z}^{(j)}) \right\rangle$,
% (3) $\textnormal{GC}(\mathbf{z}^{(i)}, \mathbf{z}^{(j)}) = \frac{\left\langle \bm{g}_{\hat{\bm\theta}}(\mathbf{z}^{(i)}), \bm{g}_{\hat{\bm\theta}}(\mathbf{z}^{(j)}) \right\rangle}{||\bm{g}_{\hat{\bm\theta}}(\mathbf{z}^{(i)})||||\bm{g}_{\hat{\bm\theta}}(\mathbf{z}^{(j)})||}$
% %\textnormal{cos}(\mathbf{g}_{\hat \theta}(\mathbf{z}^{(i)}), \mathbf{g}_{\hat\theta}(\mathbf{z}^{(j)}))$,
% (4) $\textnormal{TracIn}(\mathbf{z}^{(i)}, \mathbf{z}^{(j)}) = \sum_{t=1}^T \eta_t \left\langle \bm{g}_{\hat\theta}(\mathbf{z}^{(i)}), \bm{g}_{\hat{\bm\theta}}(\mathbf{z}^{(j)}) \right\rangle$, 
where $\bm{H}^{-1}_{\hat{\bm\theta}}$ is the inverse hessian matrix, $T$ is the number of epochs, and $\eta_t$ and $\bm \theta^{(t)}$ are the learning rate and optimal weights, at checkpoint $t$-th, respectively. $||\cdot||$ denote the Euclidean norm, $\langle \cdot,\cdot \rangle$ denote the inner product.
\subsection{Other Approaches}
% \paragraph{Other approaches.}
The rule-based approach~\cite{chu2013holistic} and statistics-based approach~\cite{huang2018auto} are commonly used for structured data such as tabular data. These methods are not suitable for deep learning, as they assume convexity in the model, and the rules in many large-scale datasets are not easy to find and describe. \citet{bahri2020deep} use $k$NN on the logit of a DNN to filter out noisy samples. \citet{gao2016resistance, reeve2019fast} focus on examining the resilience of $k$NN to random noise. \citet{wu2020topological} proposed TopoFilter that filters noisy by utilizing the $k$NN and Euclidean distance between pre-logits. \citet{chong-etal-2022-detecting} uses the negative log probability of the assigned classes, akin to the Self-Confidence method. NoiseGPT~\cite{wang2024noisegpt} leverages Multimodal LLMs to measure the ``probability curvature'' of data points. This curvature differs between benign and noisy samples. NoiseGPT then utilizes the In-Context Discrepancy to suggest or select corrected labels, serving as a knowledge expert for label noise detection and rectification. Unlike these prior works, this work proposes a post hoc, model-agnostic similarity-based approach for both detection and rectification, with theoretical justification under diverse, human-oriented noise.
\section{Methodology}
\subsection{Motivation}
\label{sec:motivation}
As a first start, we design a control experiment that randomly selects data points from the dataset and replaces their labels with random labels sampled from the output space.
\begin{figure}[!ht] 
\centering 
\subfigure[Cosine similarity]{\includegraphics[width=0.23\textwidth]{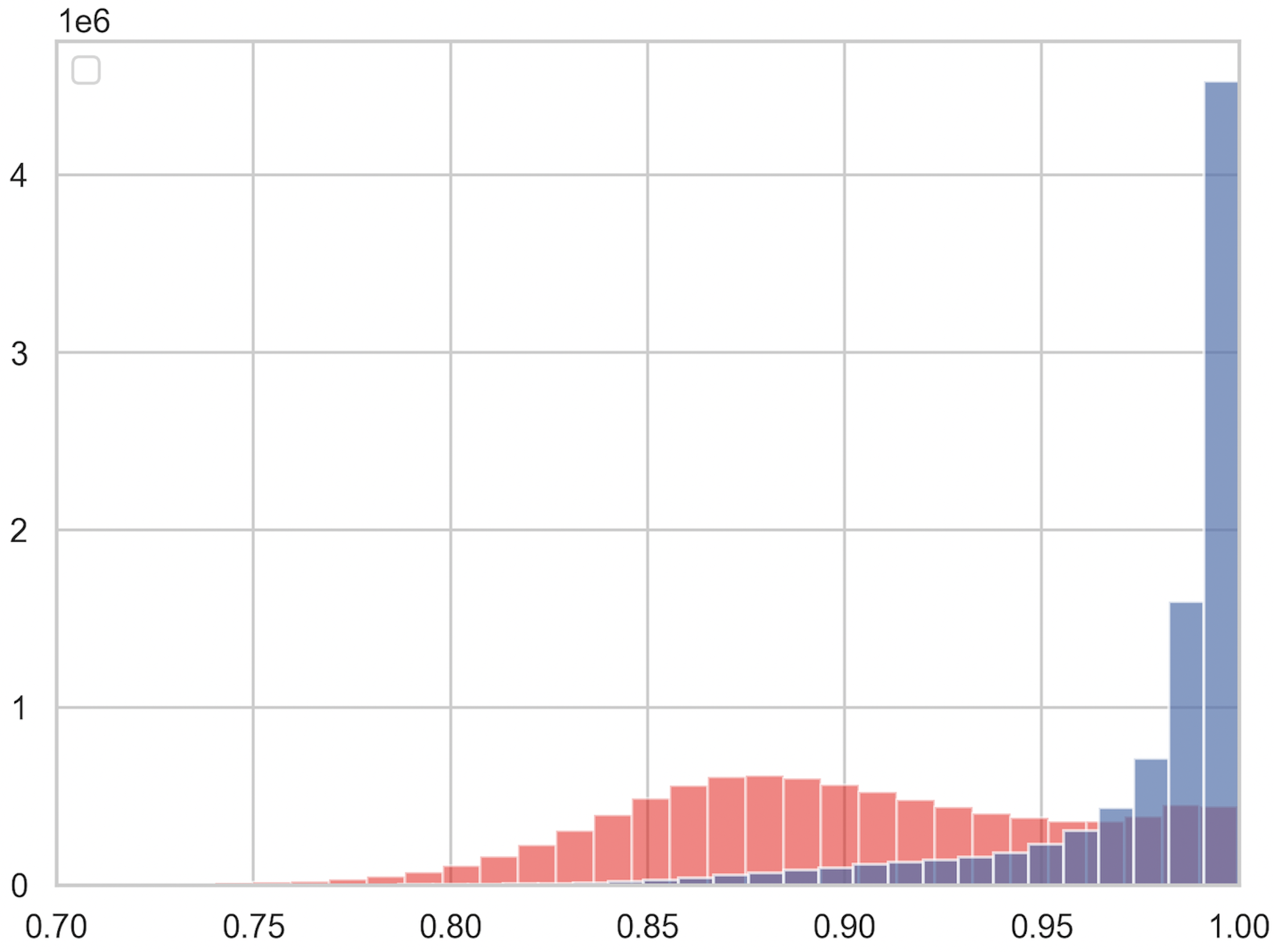}
}
\subfigure[Dot product]{\includegraphics[width=0.22\textwidth]{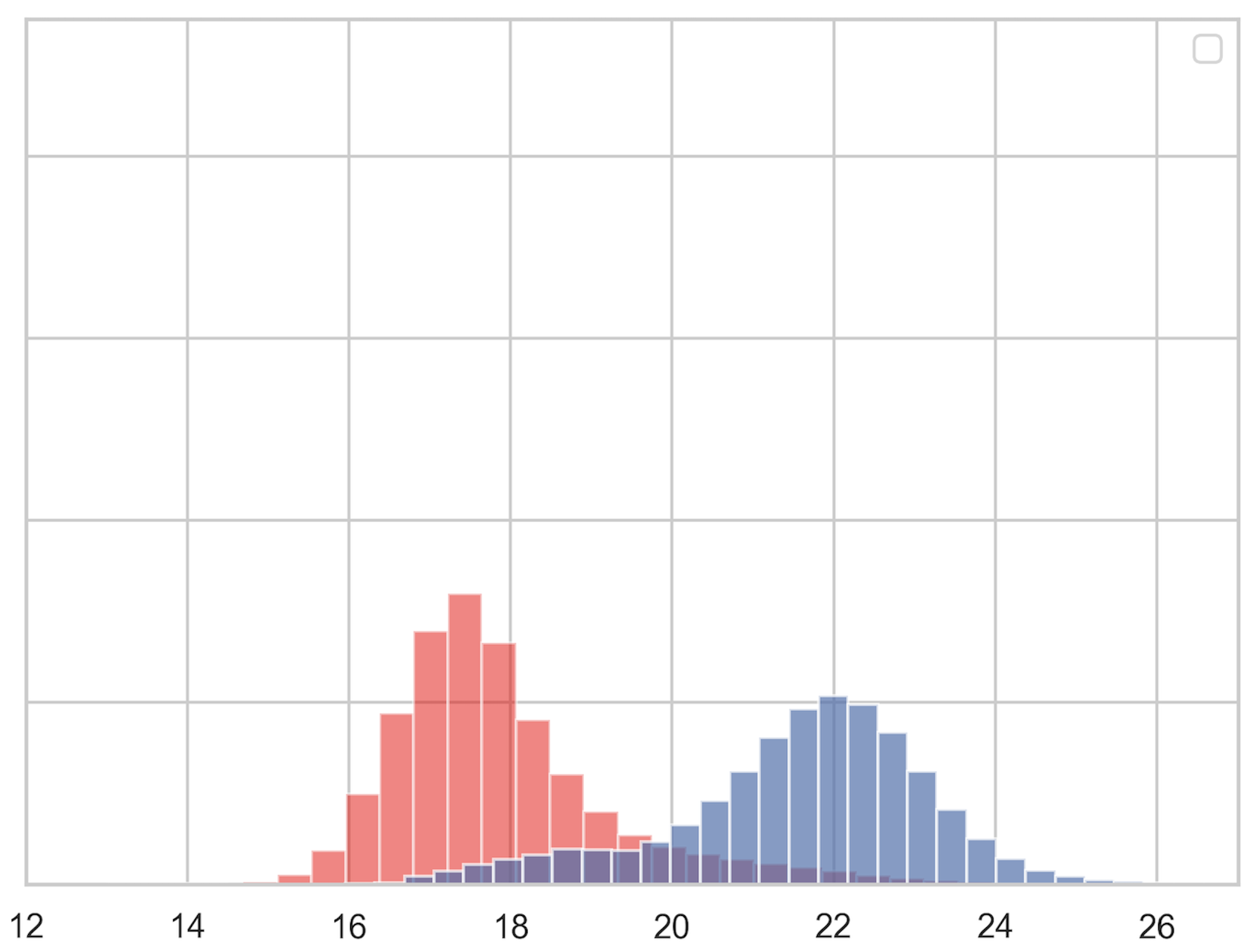}
}
\caption{Distribution of (a) Cosine similarity and (b) Dot product over IMDB~\cite{imdb} with $10\%$ noise. \textcolor{blue}{Blue} bars represent the similarity between mislabeled data points and their true class data points, \textcolor{red}{red} bars represent the similarity between mislabeled data points and other class data points. Features are obtained from a trained BERT.}
\label{fig1}
\end{figure}
We then train a deep network using gradient descent on these altered datasets to measure how noisy data points behave on other data points in the penultimate latent space, \textit{i.e.,} the representation space before softmax. We then employ the cosine similarity and the dot product similarity to compute the similarity between the mislabeled data points and data points from other classes. 
Surprisingly, as an illustrative example in Fig.~\ref{fig1}, we observed that the similarity between the mislabeled data points and their true class data point penultimate-layer representations is often higher than that of other class data points. 
We hypothesize that in overparameterized models, when the noise rate is sufficiently small, around $5-10\%$---which is a realistic assumption in practice---the model tends to memorize the noise while approaching the optimum, causing the gradient contribution of the noisy samples to diminish. Consequently, as the model converges, the influence of noise on the decision boundary becomes negligible. To support this observation, we further provide a theoretical analysis in Section~\ref{theoretical_analysis}.
%Notably, the occurrence frequency of class labels within a tightly similar cluster proves informative to detect and rectify label errors.
\subsection{Theoretical Analysis}
\label{theoretical_analysis}
Let us consider a DNN that receives input $\mathbf{x}\in \mathbb{R}^{d}$, where $d$ is the input dimension. Let $\delta$ be the softmax activation function, $\mathbf{b}\in \mathbb{R}^{N}$ be the bias term. Given a function $\phi_{\bm W}$ parameterized by $\bm W \in \mathbb{R}^{N\times d}$. For two inputs $\mathbf{x}^{(i)}$ and $\mathbf{x}^{(j)}$ we can obtain two output vectors $\phi_{\bm W}(\mathbf{x}^{(i)})$ and $\phi_{\bm W}(\mathbf{x}^{(j)})$ from $\phi_{\bm W}$, respectively. As seen by a DNN, we can measure the influence between $\mathbf{x}^{(i)}$ and $\mathbf{x}^{(j)}$ by quantifying how much $\phi_{\bm W}(\mathbf{x}^{(i)})$ change would change $\phi_{\bm W}(\mathbf{x}^{(j)})$ as well. If $\mathbf{x}^{(i)}$ and $\mathbf{x}^{(j)}$ have high similarity, then $\mathbf{x}^{(i)}$ have high influence on $\mathbf{x}^{(j)}$ and changing $\phi_{\bm W}(\mathbf{x}^{(i)})$ have large effect on $\phi_{\bm W}(\mathbf{x}^{(j)})$. Otherwise, if they have low similarity, then $\mathbf{x}^{(i)}$ have low influence on $\mathbf{x}^{(j)}$ and changing $\phi_{\bm W}(\mathbf{x}^{(i)})$ have small effect on $\phi_{\bm W}(\mathbf{x}^{(j)})$. To measure the similarity between data points, we employ a symmetric kernel proposed by~\citet{charpiat2019input}: The Inner Product kernel
\begin{align}
    \mathcal{K}(\mathbf{x}^{(i)},\mathbf{x}^{(j)}) =  \left\langle\nabla_{\bm W}\phi_{\bm W}(\mathbf{x}^{(i)}), \nabla_{\bm W}\phi_{\bm W}(\mathbf{x}^{(j)})\right\rangle, \label{eq1}
\end{align}
We choice $\phi$ is the Cross Entropy between softmax activation output $\hat{\mathbf{y}} = \delta(\bm W\mathbf{x}+\mathbf{b})$ and true distribution $\mathbf{y}$. We have $\phi_{\bm W}(\mathbf{x}) = \ell(\hat{\mathbf{y}}, \mathbf{y}; \bm W)$. For simplicity, we remove the bias term. Let denote $\mathbf{u} = \bm W\mathbf{x}$ and $\ell(\hat{\mathbf{y}}, \mathbf{y}) = \ell(\hat{\mathbf{y}}, \mathbf{y}; W)$. Using the chain rule, we obtain:
\begin{align}
    \nabla_{\bm W}\ell(\hat{\mathbf{y}}, \mathbf{y})&=\textnormal{vec}\left[\frac{\partial \ell(\hat{\mathbf{y}}, \mathbf{y})}{\partial \bm W}\right]\nonumber= \textnormal{vec}\left[\frac{\partial \ell(\hat{\mathbf{y}}, \mathbf{y})}{\partial \mathbf{u}} \frac{\partial \mathbf{u}}{\partial \bm W}\right] \nonumber\\
    & = \nabla_{\mathbf{u}}\ell(\hat{\mathbf{y}}, \mathbf{y}) \mathbf{x}^{\top}, \label{eq2}
\end{align}
where vec$\left[\frac{\partial \ell(\hat{\mathbf{y}}, \mathbf{y})}{\partial \bm W}\right]$ is the vectorization of the derivative of the loss $\ell$ w.r.t. $\bm W$.
The partial $\frac{\partial \ell(\hat{\mathbf{y}}, \mathbf{y})}{\partial \mathbf{u}}$ is 
\begin{align}
    \frac{\partial \ell(\hat{\mathbf{y}}, \mathbf{y})}{\partial \mathbf{u}} = \frac{\partial \ell(\hat{\mathbf{y}}, \mathbf{y})}{\partial \hat{\mathbf{y}}} \frac{\partial \hat{\mathbf{y}}}{\partial \mathbf{u}}
    \label{eq3}
\end{align}
The first term on the right-hand side of Eqn.~\ref{eq3} is the partial derivatives of the loss w.r.t. the predicted output $\hat{\mathbf{y}}$. Regarding to the fact: $\mathbf{y}$ is the one-hot vector present label $k$ has element $y_k = 1$ and $y_i = 0$ if $i\neq k$. We have:
\begin{align}
    \frac{\partial \ell(\hat{\mathbf{y}}, \mathbf{y})}{\partial \hat{\mathbf{y}}} &= \begin{bmatrix}
        \frac{\partial \ell (\hat{\mathbf{y}}, \mathbf{y})}{\partial \hat{y}_1} \cdots \frac{\partial \ell (\hat{\mathbf{y}}, \mathbf{y})}{\partial \hat{y}_k} \cdots \frac{\partial \ell (\hat{\mathbf{y}}, \mathbf{y})}{\partial \hat{y}_N}
        \end{bmatrix}\nonumber\\
    & = \begin{bmatrix}
            0 \cdots \frac{1}{\hat{y}_k} \cdots 0
        \end{bmatrix}\label{eq4}
\end{align}
The second term is the matrix that comprises partial derivatives of the predicted output $\hat{\mathbf{y}}$ w.r.t. $\mathbf{u}$. Regarding the fact: 

\begin{align}
    \frac{\partial \hat{\mathbf{y}}}{\partial \mathbf{u}}=\begin{bmatrix}
\frac{\partial \hat{y}_1}{\partial u_1} & \frac{\partial \hat{y}_1}{\partial u_2} & \cdots & \frac{\partial \hat{y}_1}{\partial u_{N}} \\
\vdots & \vdots & \vdots & \vdots \\
\frac{\partial \hat{y}_k}{\partial u_1} & \frac{\partial \hat{y}_k}{\partial u_2} & \cdots & \frac{\partial \hat{y}_k}{\partial u_{N}} \\
\vdots & \vdots & \vdots & \vdots \\
\frac{\partial \hat{y}_{N}}{\partial u_1} & \frac{\partial \hat{y}_{N}}{\partial u_2} & \cdots & \frac{\partial \hat{y}_{N}}{\partial u_{N}} \\
\end{bmatrix}\label{eq5}
\end{align}
Substitute Eqn.~\ref{eq5} and Eqn.~\ref{eq4} into Eqn.~\ref{eq3}, we get 
\begin{align}
    &\frac{\partial \ell(\hat{\mathbf{y}}, \mathbf{y})}{\partial \mathbf{u}} = \nonumber \\
    &\begin{bmatrix}
\frac{\partial \ell(\hat{\mathbf{y}}, \mathbf{y})}{\partial \hat{y}_k} \frac{\partial \hat{y}_k}{\partial u_1} & \cdots & \frac{\partial \ell(\hat{\mathbf{y}}, \mathbf{y})}{\partial \hat{y}_k} \frac{\partial \hat{y}_k}{\partial u_k} & \cdots & \frac{\partial \ell(\hat{\mathbf{y}}, \mathbf{y})}{\partial \hat{y}_k} \frac{\partial \hat{y}_k}{\partial u_N}
\end{bmatrix}\label{eq6}
\end{align}
Given a softmax activation function $\delta$ for class $k$: 
Case 1: $i = k$, we compute the derivative of softmax output $\hat{y}_k$ w.r.t. $u_k$:
\begin{align}
    \frac{\partial \hat{y}_k}{\partial u_k} &= \frac{\partial}{\partial u_k} \left( \frac{e^{u_k}}{\sum_{i=1}^{N}e^{u_i}}\right)\nonumber \\ 
    & = \frac{e^{u_k} \left(\sum_{i=1}^{N}e^{u_i}\right) - e^{u_k}\cdot e^{u_k}}{\left(\sum_{i=1}^{N}e^{u_i}\right)^2} \nonumber\\ 
    & = \hat{y}_k (1-\hat{y}_k)\label{eq7}
\end{align}
Case 2: $i \neq k$, we compute the derivative of softmax output $\hat{y}_k$ w.r.t. $u_i$:
\begin{align}
    \frac{\partial \hat{y}_k}{\partial u_i} = \frac{\partial}{\partial u_i} \left(\frac{e^{u_k}}{\sum_{i=1}^{N}e^{u_i}}\right)\label{eq8}
\end{align}
Using the chain rule, we obtain:
\begin{align}
    \frac{\partial \hat{y}_k}{\partial u_i} = - \frac{e^{u_k} \cdot e^{u_i}}{\left(\sum_{i=1}^{N}e^{u_i}\right)^2} = -\hat{y}_k \hat{y}_i\label{eq9}
\end{align}
Substitute Eqn.~\ref{eq7} and Eqn.~\ref{eq9} into Eqn.~\ref{eq6}, we get a column vector:
\begin{align}
 \nabla_{\mathbf{u}}\ell(\hat{\mathbf{y}}, \mathbf{y}) = \begin{bmatrix}
-\hat{y}_1 
\cdots 
1 - \hat{y}_k 
\cdots
-\hat{y}_N
\end{bmatrix}^{\top}\label{eq11}
\end{align}
The inner product kernel in Eqn.~\ref{eq1} becomes:
\begin{align}
    \mathcal{K}(\mathbf{x}^{(i)},\mathbf{x}^{(j)}) = &\overbrace{\nabla_{\mathbf{u}}\ell(\hat{\mathbf{y}}^{(i)}, \mathbf{y}^{(i)})^{\top}\nabla_{\mathbf{u}}\ell(\hat{\mathbf{y}}^{(j)}, \mathbf{y}^{(j)})}^{\mathbf{G}^{(ij)}} \nonumber \\
&\cdot (\mathbf{x}^{(j)\top} \mathbf{x}^{(i)})\label{eq11}
\end{align}
We denote $\mathbf{G}^{(ij)}$ as the product of two gradients of the loss at $\mathbf{x}^{(i)}$ and $\mathbf{x}^{(j)}$. Suppose input $\mathbf{x}^{(i)}$ and input $\mathbf{x}^{(j)}$ have label $k$ and $k'$ corresponding. We have
\begin{align}
    \mathbf{G}^{(ij)} &= \begin{bmatrix}-\hat{y}^{(i)}_1 \cdots 1-\hat{y}^{(i)}_k \cdots-\hat{y}^{(i)}_N
\end{bmatrix}  \begin{bmatrix}
-\hat{y}^{(j)}_1 \\
\vdots \\
1 - \hat{y}^{(j)}_{k'} \\
\vdots \\
-\hat{y}^{(j)}_N
\end{bmatrix} \label{eq12}
\end{align}
We consider 2 cases:\\
If $k = k'$:
\begin{align}
\mathbf{G}^{(ij)}_{k =k'} &= \hat{y}^{(i)}_1\hat{y}^{(j)}_1 + \cdots + (1-\hat{y}^{(i)}_k)(1-\hat{y}^{(j)}_k) \nonumber\\&+ \cdots +  \hat{y}^{(i)}_N\hat{y}^{(j)}_N \nonumber\\
&=(1 - \hat{y}^{(i)}_k) (1 - \hat{y}^{(j)}_k) + \sum_{n = 1, n \neq k}^{N} \hat{y}^{(i)}_n \hat{y}^{(j)}_n\label{eq13}
\end{align}
If $k \neq k'$:
\begin{align}
\mathbf{G}^{(ij)}_{k \neq {k'}} &= \hat{y}^{(i)}_1\hat{y}^{(j)}_1 + \cdots + (1-\hat{y}^{(i)}_k)(-\hat{y}^{(j)}_k) +\nonumber \\ &(1-\hat{y}^{(j)}_{k'})(-\hat{y}^{(i)}_{k'})+ \cdots +  \hat{y}^{(i)}_N\hat{y}^{(j)}_N \nonumber\\
&=\hat{y}^{(j)}_k(\hat{y}^{(i)}_k - 1) + \hat{y}^{(i)}_{k'}(\hat{y}^{(j)}_{k'} - 1) \nonumber\\&+ \sum_{n = 1, n \neq k, n\neq k'}^{N} \hat{y}^{(i)}_n \hat{y}^{(j)}_n\label{eq14}
\end{align}
During the training process, the model is more confident about the labels of data points, indicating the value of $\hat{y}^{(i)}_k$ and $\hat{y}^{(j)}_{k'}$ being closer to $1$. Assume a well-trained model, and $\hat{y}^{(i)}_k \approx \hat{y}^{(j)}_{k'} = \alpha$; $\hat{y}^{(i)}_n =\hat{y}^{(j)}_n \approx \epsilon = \frac{1-\alpha}{N-1} $. ($n\neq k$ and $n \neq k'$). Substitute these values into Eqn.~\ref{eq13} and Eqn.~\ref{eq14}, we get:
\begin{align}
    \mathbf{G}^{(ij)}_{k=k'} &\approx (1-\alpha)^2 +\epsilon^2(N-1)\label{eq15}\\
    \mathbf{G}^{(ij)}_{k\neq {k'}} &\approx -\frac{N(1-\alpha)^2}{(N-1)^2} = -\epsilon^2 N\label{eq16}
\end{align}
%As assumsed, if the network is well trained: $\hat{y}^{(i)}_k$ and $\hat{y}^{(j)}_{k'}$ approximately $1$, then 
As $N$ become very large with deep learning dataset and $\epsilon$ small, the magnitude of $\mathbf{G}^{(ij)}_{k\neq {k'}}$ is close to $0$ for $k \neq k'$. That means data points in different classes tend to be pushed into different orthogonal subspaces. Let's consider the  the magnitude of $\mathbf{G}^{(ij)}_{k=k'}$ and $\mathbf{G}^{(ij)}_{k \neq {k'}}$, divide $|\mathbf{G}^{(ij)}_{k=k'}|$ by $|\mathbf{G}^{(ij)}_{k\neq{k'}}|$, we get:
\begin{align}
    \frac{\mathcal{K}^{(ij)}_{k=k'}}{\mathcal{K}^{(ij)}_{k\neq {k'}}}=\frac{|\mathbf{G}^{(ij)}_{k=k'}|}{|\mathbf{G}^{(ij)}_{k \neq {k'}}|} &\approx \frac{|(1-\alpha)^2 + \epsilon^2(N-1)|}{|-\epsilon^2N|}\nonumber \\
    & \approx \frac{\epsilon^2(N-1)^2 + \epsilon^2(N-1)}{\epsilon^2N}\nonumber \\
    & \approx N - 1 \\
    \mathcal{K}^{(ij)}_{k=k'} & \approx (N - 1) \mathcal{K}^{(ij)}_{k\neq {k'}}  \label{eq22}
\end{align}
Here, the kernel $\mathcal{K}^{(ij)}_{k=k'}$ ($\mathcal{K}^{(ij)}_{k\neq k'}$) represents the similarity between $\mathbf{x}^{(i)}$ and $\mathbf{x}^{(j)}$ when they share the same label (different labels). The model tends to overfit to the noisy label, resulting in higher similarity with the mislabeled class than the true class. This analysis explains why mislabeled data points are often more similar to true class data points than data points in other classes. Note that the cosine similarity is a normalized inner product; hence, the analysis remains valid for cosine similarity. Last, in datasets with a larger number of classes (larger $N$), data points belonging to the same class tend to lie closer together within their respective subspaces. Consequently, similarity-based methods are expected to perform better in such settings. We provide empirical evidence supporting this claim in Section~\ref{sec:result}.
\subsection{Algorithm}
\label{sec:method}
Motivated by the analysis, we propose a post-hoc, model-agnostic approach, as detailed in Algorithm~\ref{alg:NNerr}. It requires a small auxiliary dataset $\mathcal{D}_{\textnormal{aux}}$, a similarity measure $\sigma (\cdot, \cdot)$. We denote $\phi^{(i)}$ and $\phi^{(j)}$ be the penultimate feature representations of $\mathbf{z}^{(i)}$ and $\mathbf{z}^{(j)}$ obtained from the trained model $f_{\hat{\theta}}$ respectively. We employ two primary similarity measures: Dot product ($\textnormal{DOT} = \left\langle \phi^{(i)}, \phi^{(j)}\right\rangle$) and Cosine similarity ($\textnormal{COS} = \frac{\left\langle \phi^{(i)}, \phi^{(j)}\right\rangle}{||\phi^{(i)}||||\phi^{(j)}||}$). We denote $\mathcal{S}(\mathcal{D}_{\textnormal{aux}}, \mathbf{z}^{(i)})$ as $k$ most similar to $\mathbf{z}^{(i)}$ in $\mathcal{D}_{\textnormal{aux}}$.
Given a noisy dataset $\mathcal{D}$, for each data point $\mathbf{z}^{(i)} \in \mathcal{D}$ with label $\mathbf{y}^{(i)}$, our algorithm finds $\mathcal{S}(\mathcal{D}_{\textnormal{aux}}, \mathbf{z}^{(i)})$ such that every data point in $\mathcal{D}$ but not in $\mathcal{S}(\mathcal{D}_{\textnormal{aux}}, \mathbf{z}^{(i)})$ is at most similar to $\mathbf{z}^{(i)}$ as the least similar point in $\mathcal{S}(\mathcal{D}_{\textnormal{aux}}, \mathbf{z}^{(i)})$ (line~\ref{al:line6}). 
\begin{algorithm}[t!]
\caption{Noise Detection and Rectification}\label{alg:NNerr}
\begin{algorithmic}[1]
\Require \\
$\mathcal{D} = \left\lbrace \mathbf{z}^{(i)}  \right\rbrace_{i=1}^n 
$: a noisy dataset\\
$\mathcal{D}_{\textnormal{aux}} = \left\lbrace \mathbf{z}^{(j)}  \right\rbrace_{j=1}^{m}$: an auxiliary dataset.\\
$\sigma(\cdot, \cdot)$: a similarity measure.\\
$k$: number of most similar data points.
\Ensure noisy data points in $\mathcal{D}$ are rectified.
    % \Statex \algcomment{/* \textbf{Error Detection} */}
    \For{$\mathbf{z}^{(i)} \in \mathcal{D}$} 
        \State $\mathcal{S}(\mathcal{D}_{\textnormal{aux}}, \mathbf{z}^{(i)}) = \{\mathbf{z}^{(j)} \in \mathcal{D}_{\textnormal{aux}}\}$ \label{al:line6}
        
        \noindent s.t. $|\mathcal{S}(\mathcal{D}_{\textnormal{aux}}, \mathbf{z}^{(i)})| = k$, and
        $$\sigma(\mathbf{z}^{(i)}, \mathbf{z}^{(j)}) \geq
         \max_{\mathbf{z}'^{(i)} \in \mathcal{D}\backslash\mathcal{S}(\mathcal{D}_{\textnormal{aux}}, \mathbf{z}^{(i)})} \sigma(\mathbf{z}^{(i)}, \mathbf{z}'^{(i)})$$
         \State $\mathbf{s}^{(i)} = \frac{1}{k}\sum_{\mathbf{z}^{(j)}\in \mathcal{S}(\mathcal{D}_{\textnormal{aux}}, \mathbf{z}^{(i)})} \mathbb{I}(\mathbf{y}^{(j)} = \mathbf{y}^{(i)})$ \label{al:line7}
        %\State $s^{(i)} \gets \frac{count(y^{(i)}, \psi^{(i)})}{k}$
    \EndFor
    \State $\mathcal{D}^{\uparrow} = \textnormal{sort}(\mathcal{D}, \textnormal{key}= \mathbf{s}, \textnormal{ascending}=\textnormal{True})$ \label{al:line9}
    \For{$\mathbf{z}^{(i)} \in \mathcal{D}^{\uparrow}_{:p}$}
        \State $\mathbf{z}^{(i)} = \left(\mathbf{x}^{(i)}, \textsc{Mode}(\mathcal{S}(\mathcal{D}_{\textnormal{aux}}, \mathbf{z}^{(i)}))\right)$ \label{line10}
    \EndFor\\
    \Return $\mathcal{D}$

\end{algorithmic}
\end{algorithm}
We define a scoring function that return $\mathbf{s}^{(i)}$---the probability of occurrence of label $\mathbf{y}^{(i)}$ in $\mathcal{S}(\mathcal{D}_{\textnormal{aux}}, \mathbf{z}^{(i)})$ (line~\ref{al:line7}). The indicator $\mathbb{I}(\cdot)$ returns $1$ if the condition holds. The lower $\mathbf{s}$ is, the more likely a label noise. We sort the data points in $\mathcal{D}$ in ascending order of $\mathbf{s}$ and obtain the sorted $\mathcal{D}^\uparrow$ (line~\ref{al:line9}). We select the first $p\%$ samples of ranked set $\mathcal{D}^{\uparrow}$ denoted as $\mathcal{D}^{\uparrow}_{:p}$ and define a class decision rule $\textsc{Mode}(\cdot)$ that selects the label in $\mathcal{S}(\mathcal{D}_{\textnormal{aux}}, \mathbf{z}^{(i)})$ has the highest probability and greater than threshold $\tau$. Otherwise, the label of $\mathbf{z}^{(i)}$ remains unchanged (line~\ref{line10}).

\section{Experiment}
\subsection{Dataset and Model}
We use \textbf{Snippets \textnormal{\cite{snippet}}} and \textbf{IMDB \textnormal{\cite{imdb}}} for evaluation. Snippets is a dataset of web search snippets retrieved
from Google Search with 8 domains, including business, computers, culture-arts-entertainment, education-science, engineering, health, politics-society, and sports. The training data and testing data include $10,060$ and $2,280$ snippets, respectively. For validation purposes, we randomly split the original training data into train/validation with the ratio of $8048/2012$. 
%The dataset can be found at \url{http://jwebpro.sourceforge.net/data-web-snippets.tar.gz}.
%\paragraph{Ohsumed \textnormal{\cite{hersh1994ohsumed}}} The dataset includes medical abstracts from the MeSH categories of the year 1991. The
%specific task was to categorize the 23 cardiovascular diseases. The original dataset could be found at \url{http://disi.unitn.it/moschitti/corpora.htm}
% \paragraph{SST-5 \textnormal{\cite{sst5}}} Stanford Sentiment Treebank is a corpus with fully labeled parse
% trees that allows for a complete analysis of the
% compositional effects of sentiment in language. It
% consists of 11,855 single sentences extracted from
% movie reviews. We use this dataset for the sentiment analysis problem. We use the corpus with 5 labels. The labels are negative, somewhat negative, neutral, somewhat positive, and positive. Train/test/valid: ???. The dataset could be found at \url{https://deepai.org/dataset/stanford-sentiment-treebank}.
%\paragraph{TweetEval \textnormal{\cite{barbieri2020tweeteval}}} consists of $7$ heterogeneous tasks in Twitter, all framed as multi-class tweet classification. All tasks have been unified into the same benchmark. We use the Emotion Recognition task~\cite{mohammad2018semeval}. The dataset consists of $5053$ tweets with $4$ labels: anger, joy, sadness, and optimism. Train/test/validation: $3257/1422/374$ The dataset can be found at \url{https://github.com/cardiffnlp/tweeteval}.
IMDB is the most common benchmark for sentiment analysis tasks. IMDB includes $50,000$ reviews from the Internet Movie Database website with original $25,000$ negative and $25,000$ positive reviews. For validation purposes, we randomly split into training, validation, and test sets of sizes $15,000$, $5,000$, and $25,000$. 
% The IMDB dataset can be found at~\url{https://ai.stanford.edu/~amaas/data/sentiment/}. 
We use \textbf{BERT}~\cite{bert} base uncased version, which is one of the most standard used models for language understanding tasks. We use open-weight LLMs: \textbf{Llama-3.2-1B, Llama-3.2-3B, Llama-3.1-8B}~\cite{llama3modelcard} for the zero-shot detection scenarios in Subsection~\ref{zero_shot}. Due to space constraints, we present the main results and include some additional results in the Appendix upon the final decision.

% \textbf{BERT \textnormal{\cite{bert}}} stands for Bidirectional Encoder Representations from Transformers. BERT is one of the most standard used pre-trained model for language understanding tasks. In all settings, we use BERT base uncased version.

\subsection{Experimental Setup}
\label{expdetail}
\begin{figure*}[t]
\centering
\subfigure[$5\%$ random noise]{
    \includegraphics[width=0.3\textwidth]{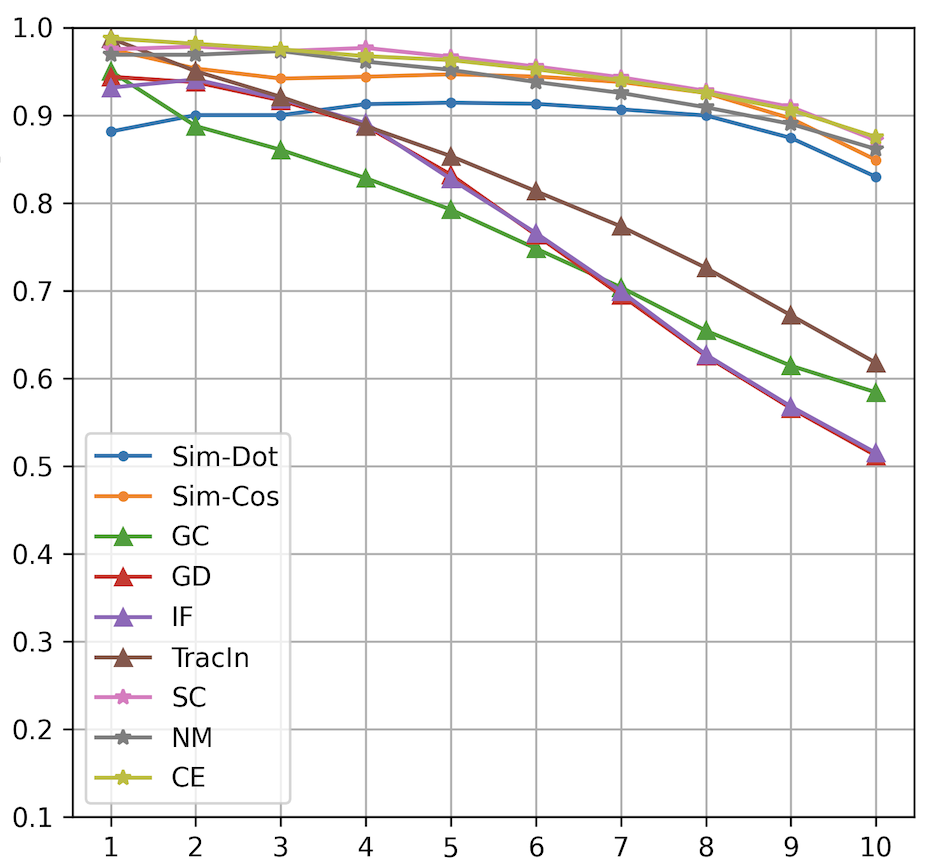}
    \label{fig2a}
}
\subfigure[$10\%$ ambiguity noise]{
    \includegraphics[width=0.3\textwidth]{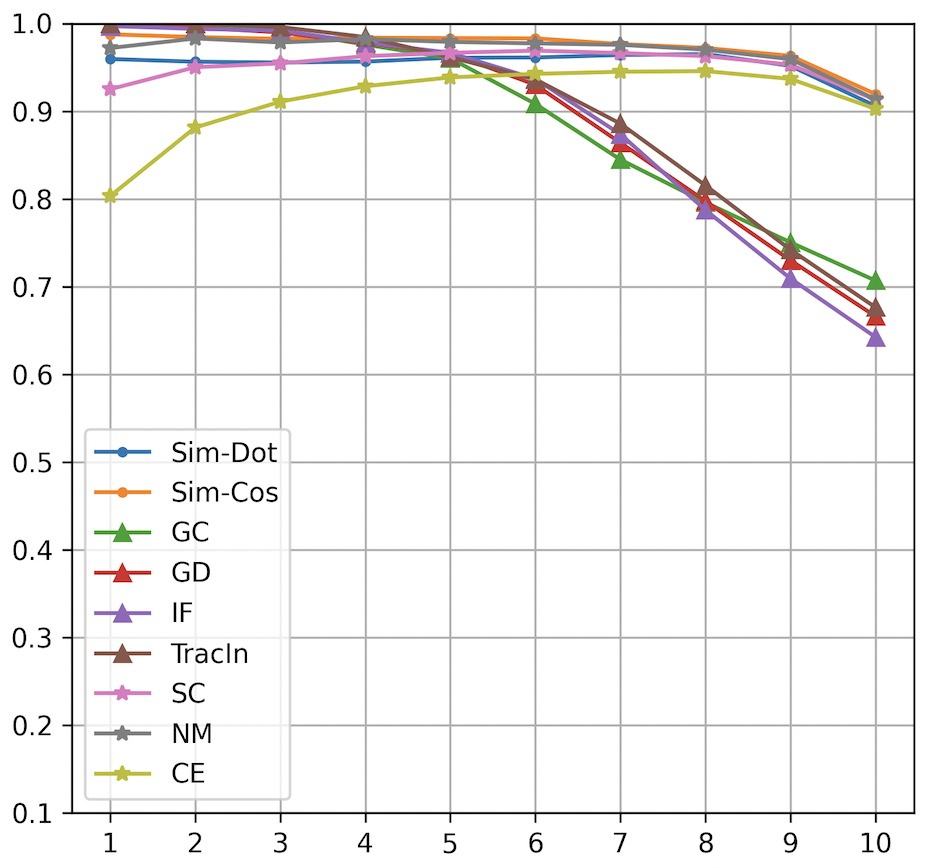}
    \label{fig2b}
}
\subfigure[$20\%$ concentrated noise]{
    \includegraphics[width=0.3\textwidth]{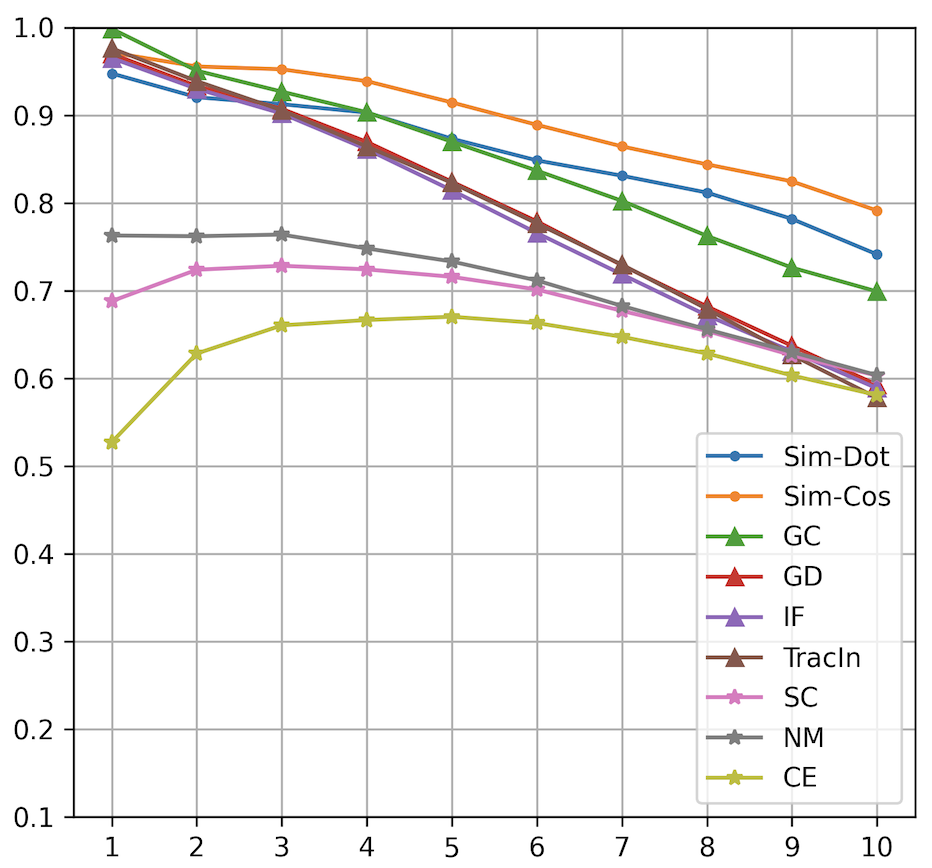}
    \label{fig2c}
}
\subfigure[$5\%$ random noise]{
    \includegraphics[width=0.3\textwidth]{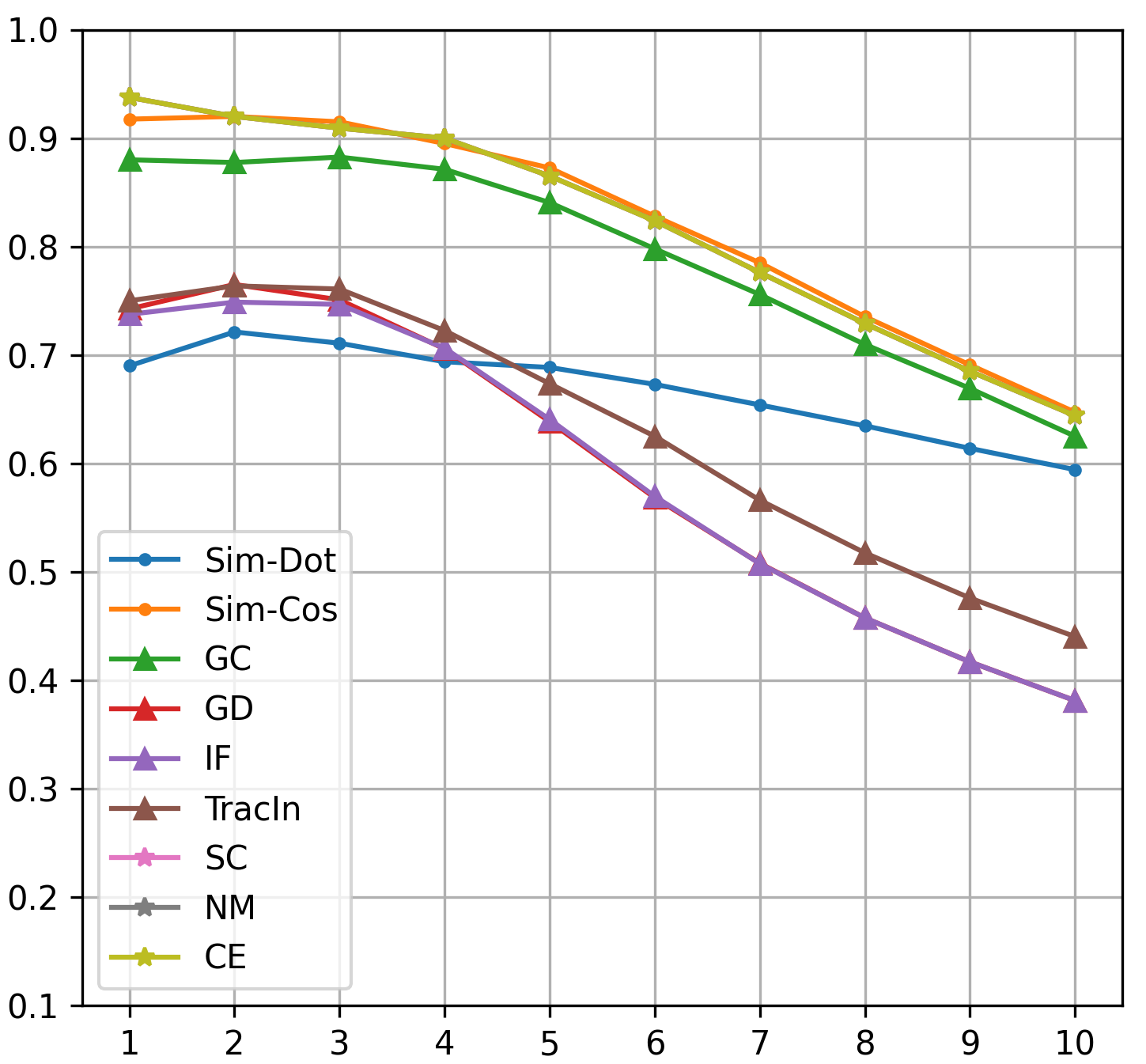}
    \label{fig2d}
}
\subfigure[$10\%$ ambiguity noise]{
    \includegraphics[width=0.3\textwidth]{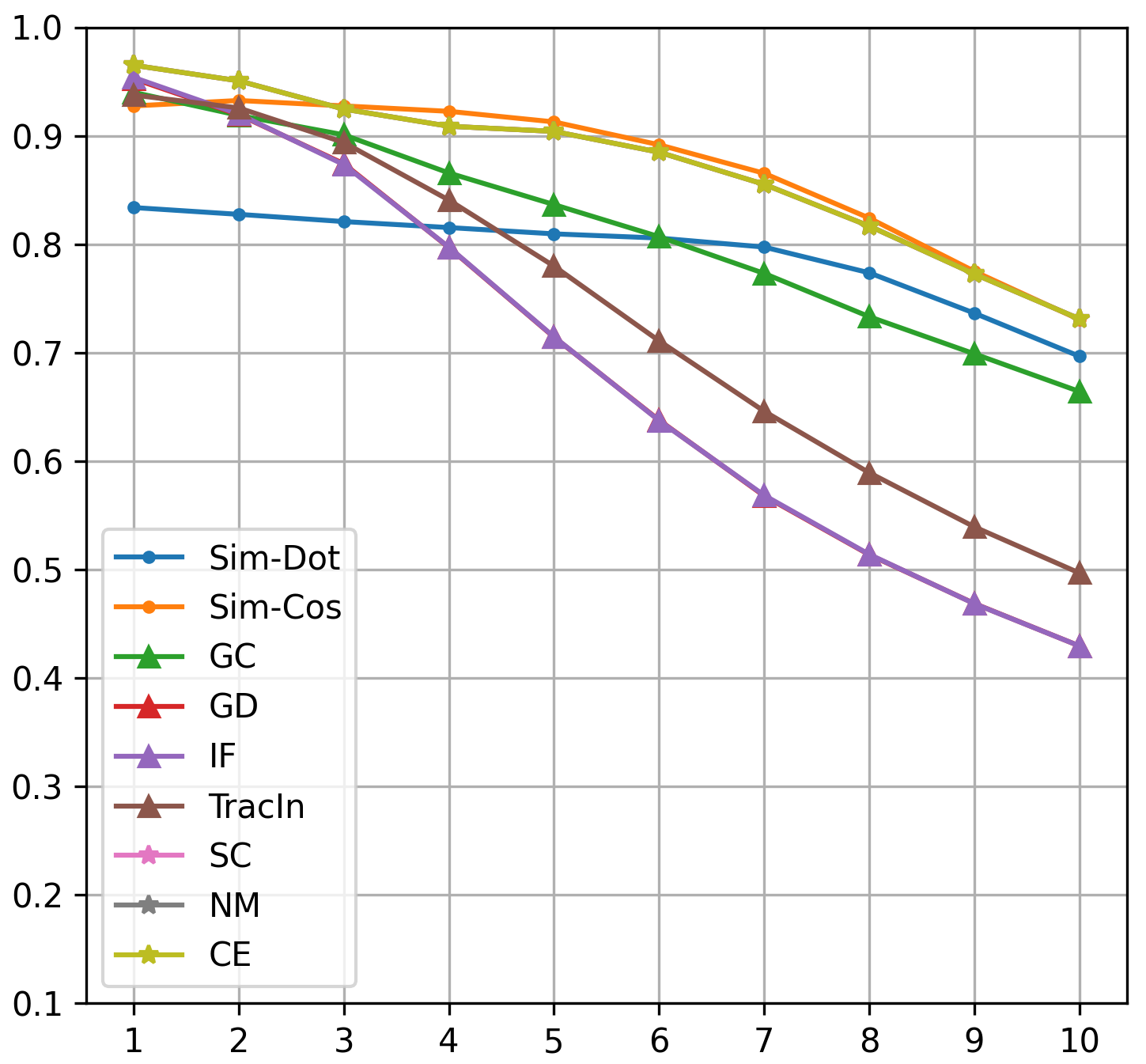}
    \label{fig2e}
}
\subfigure[$20\%$ concentrated noise]{
    \includegraphics[width=0.3\textwidth]{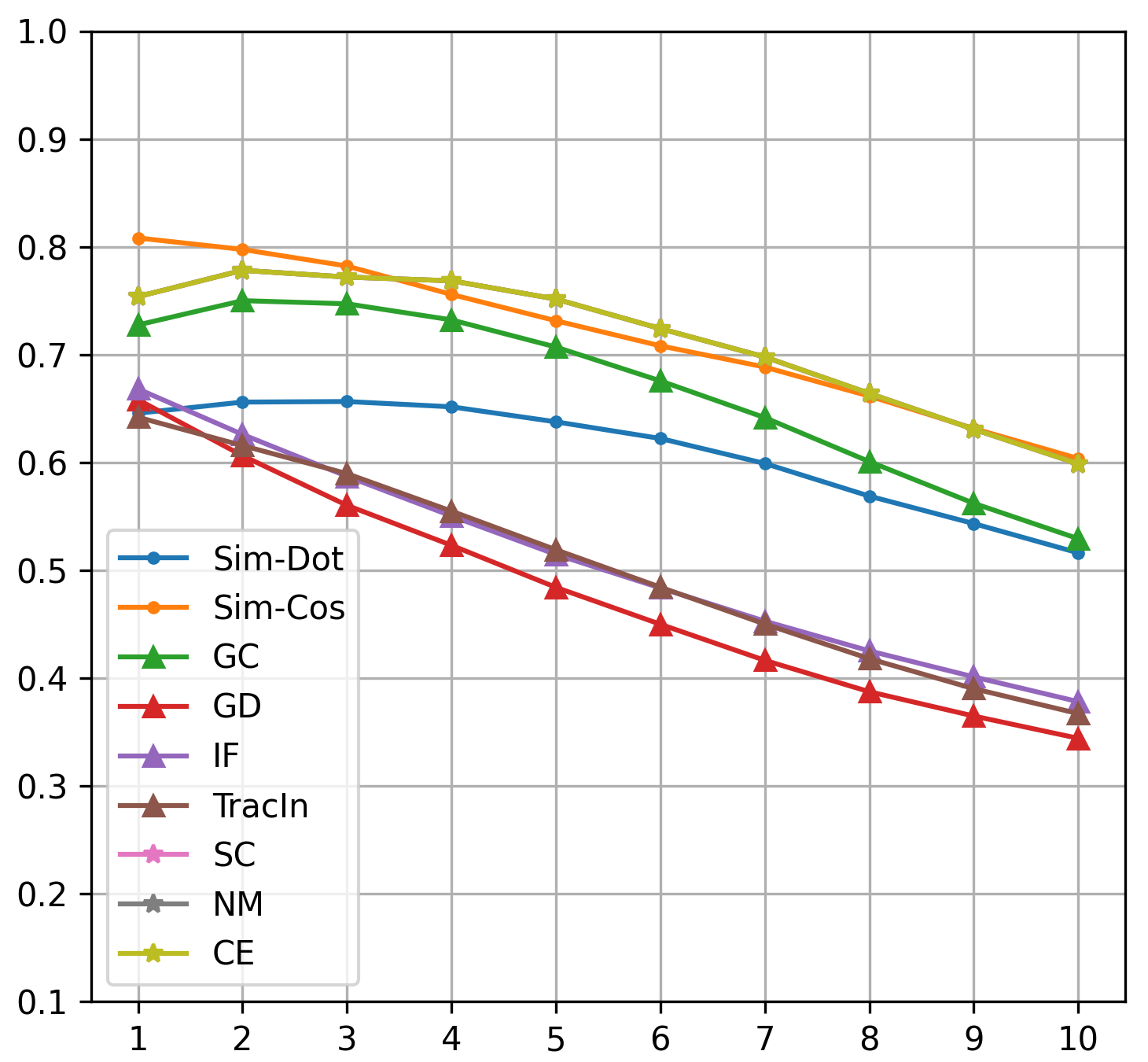}
    \label{fig2f}
}
\caption{
Noise detection accuracy of methods measured on Snippets (Fig.~\ref{fig2}a-c) and IMDB (Fig.~\ref{fig2}d-f) across various sizes and noise types. 
}
\label{fig2}
\end{figure*}

\begin{table*}[!ht]
\centering
\small
\caption{Test accuracy of model measured on Snippets after removing or rectifying top $20\%$ potential noisy samples ranked by methods. The \textbf{best}, \textcolor{blue}{improvement}, \textcolor{red}{drop}, and \underline{runner-up} are marked.}
% \label{tab:improvement}
\resizebox{\linewidth}{!}{%
    \begin{tabular}{@{}lcccccccc@{}}
    \toprule
    \multicolumn{1}{c}{\multirow{2}{*}{\textbf{Method}}} & \multicolumn{2}{c}{\textbf{Random noise}} & \multicolumn{2}{c} {\textbf{Ambiguity noise}} & \multicolumn{2}{c} {\textbf{Concentrated noise}}       
    \\ \cmidrule(lr){2-3} \cmidrule(l){4-5} \cmidrule(l){6-7} 
    \multicolumn{1}{c}{}  & Removed & Rectified  & Removed & Rectified & Removed & Rectified \\ \midrule
    Acc. (under noise)     & $88.64 $ & --- & $82.50 $ & --- & $79.38 $ & --- \\\midrule
     & \multicolumn{6}{c}{\textit{Confidence-based Approach}}
    \\
    Confident-Weighted Entropy     & $83.02${\tiny\textcolor{red}{$-5.62$}} & --- & $84.38${\tiny\textcolor{blue}{$+1.88$}} & --- & $77.50${\tiny\textcolor{red}{$-1.88$}} & --- \\
    Normalize-Margin                & $\underline{87.19}${\tiny\textcolor{red}{$-1.45$}} & --- & $\underline{87.01}${\tiny\textcolor{blue}{$+4.51$}} & --- & $77.89${\tiny\textcolor{red}{$-1.49$}} & --- \\
    Self-Confidence                 & $86.05${\tiny\textcolor{red}{$-2.59$}} & --- & $\textbf{87.36}${\tiny\textcolor{blue}{$+4.86$}} & --- & $78.02${\tiny\textcolor{red}{$-1.36$}} & --- \\ \midrule
    & \multicolumn{6}{c}{\textit{Gradient-based Approach}}
    \\
    Influence Function              & $83.24${\tiny\textcolor{red}{$-5.40$}} & --- & $81.71${\tiny\textcolor{red}{$-0.79$}} & --- & $79.51${\tiny\textcolor{blue}{$+0.13$}} & --- \\ 
    Gradient-Cosine                 & $73.81${\tiny\textcolor{red}{$-14.83$}} & --- & $83.33${\tiny\textcolor{blue}{$+0.83$}} & --- & $\textbf{82.23}${\tiny\textcolor{blue}{$+2.85$}}& --- \\
    Gradient-Dot                    & $86.53${\tiny\textcolor{red}{$-2.11$}}& --- &$ 82.01${\tiny\textcolor{red}{$-0.49$}} & --- & $78.68${\tiny\textcolor{red}{$-0.70$}} & --- \\
    TracIn                          & $85.48${\tiny\textcolor{red}{$-3.16$}}& --- & $81.71${\tiny\textcolor{red}{$-0.79$}} & --- & $76.67${\tiny\textcolor{red}{$-2.71$}}& --- \\\midrule
    & \multicolumn{6}{c}{\textit{Similarity-based Approach}}
    \\
    \textbf{Sim-Cos}                         & $\textbf{89.43}${\tiny\textcolor{blue}{$+0.79$}} & $87.85${\tiny\textcolor{red}{$-0.79$}} & $85.00${\tiny\textcolor{blue}{$+2.50$}} & $\underline{83.73}${\tiny\textcolor{blue}{$+1.23$}} & $\underline{81.53}${\tiny\textcolor{blue}{$+2.15$}} & $83.38${\tiny\textcolor{blue}{$+4.00$}} \\
    \textbf{Sim-Dot}                         & $86.71${\tiny\textcolor{red}{$-1.93$}} &$ 87.32${\tiny\textcolor{red}{$-1.32$}} & $82.98${\tiny\textcolor{blue}{$+0.48$}} & $83.95${\tiny\textcolor{blue}{$+1.45$}} &  $81.53${\tiny\textcolor{blue}{$+2.15$}}& $81.97${\tiny\textcolor{blue}{$+2.59$}}\\\bottomrule
    \end{tabular}
}
\label{tab:1}
\end{table*}

\begin{table*}[!ht]
\centering
\small
\caption{Test accuracy of model measured on IMDB after removing or rectifying top $20\%$ potential noisy samples ranked by methods. The \textbf{best}, \textcolor{blue}{improvement}, \textcolor{red}{drops}, and \underline{runner-up} are marked.}
\label{tab:2}
\resizebox{\linewidth}{!}{%
    \begin{tabular}{@{}lcccccccc@{}}
    \toprule
    \multicolumn{1}{c}{\multirow{2}{*}{\textbf{Method}}} & \multicolumn{2}{c}{\textbf{Random noise}} & \multicolumn{2}{c} {\textbf{Ambiguity noise}} & \multicolumn{2}{c} {\textbf{Concentrated noise}}       
    \\ \cmidrule(lr){2-3} \cmidrule(l){4-5} \cmidrule(l){6-7} 
    \multicolumn{1}{c}{}  & Removed & Rectified  & Removed & Rectified & Removed & Rectified \\ 
    \midrule
    Acc. (under noise)                            & $89.92$ & --- & $89.36$ & --- & $85.98$ & --- \\
    \midrule
    & \multicolumn{6}{c}{\textit{Confidence-based Approach}}
    \\
    Confidence-Weighted Entropy          & $\underline{91.00}${\tiny\textcolor{blue}{$+1.08$}} & --- & $90.73${\tiny\textcolor{blue}{$+1.37$}} & --- & $86.66${\tiny\textcolor{blue}{$+0.68$}} & --- \\
    Normalize-Margin               & $\underline{91.00}${\tiny\textcolor{blue}{$+1.08$}} & --- & $90.73${\tiny\textcolor{blue}{$+1.37$}} & --- & $86.66${\tiny\textcolor{blue}{$+0.68$}} & --- \\
    Self-Confidence                & $\underline{91.00}${\tiny\textcolor{blue}{$+1.08$}} & --- & $90.73${\tiny\textcolor{blue}{$+1.37$}} & --- & $86.66${\tiny\textcolor{blue}{$+0.68$}} & --- \\\midrule
    & \multicolumn{6}{c}{\textit{Gradient-based Approach}}
    \\
    Influence Function             & $90.95${\tiny\textcolor{blue}{$+1.03$}} & --- & $90.51${\tiny\textcolor{blue}{$+1.15$}} & --- & $87.09${\tiny\textcolor{blue}{$+1.11$}} & --- \\ 
    Gradient-Cosine                & $90.59${\tiny\textcolor{blue}{$+0.67$}} & --- & $90.93${\tiny\textcolor{blue}{$+1.57$}} & --- & $86.78${\tiny\textcolor{blue}{$+0.80$}} & --- \\
    Gradient-Dot                   & $90.43${\tiny\textcolor{blue}{$+0.51$}} & --- & $89.45${\tiny\textcolor{blue}{$+0.09$}} & --- & $82.87${\tiny\textcolor{red}{$-3.20$}} & --- \\
    TracIn                         & $90.19${\tiny\textcolor{blue}{$+0.27$}} & --- & $\underline{91.09}${\tiny\textcolor{blue}{$+1.73$}} & --- &$ 82.87${\tiny\textcolor{red}{$-3.20$}} & --- \\ \midrule
    & \multicolumn{6}{c}{\textit{Similarity-based Approach}}
    \\
    \textbf{Sim-Cos}                        & $90.70${\tiny\textcolor{blue}{$+0.78$}} & $88.73${\tiny\textcolor{red}{$-1.19$}} & $90.78${\tiny\textcolor{blue}{$+1.42$}} & $90.83${\tiny\textcolor{blue}{$+1.47$}} & $\textbf{87.76}${\tiny\textcolor{blue}{$+1.78$}} & $84.73${\tiny\textcolor{red}{$-1.25$}} \\
    \textbf{Sim-Dot}                        & $\textbf{91.49}${\tiny\textcolor{blue}{$+1.57$}} & $90.46${\tiny\textcolor{blue}{$+0.54$}} & $\textbf{91.58}${\tiny\textcolor{blue}{$+2.22$}} & $90.13${\tiny\textcolor{blue}{$+0.77$}} & $\underline{87.25}${\tiny\textcolor{blue}{$+1.27$}} & $87.46${\tiny\textcolor{blue}{$+1.48$}}\\\bottomrule
    \end{tabular}
}
\end{table*}

\noindent\textbf{Modeling realistic noise.} We construct three realistic, human-originated types of noise: 

\noindent (1) \textbf{\underline{Random noise}}: we randomly select data points and change the label to a different class.

\noindent (2) \textbf{\underline{Systematic ambiguity noise}}: we establish a rule $r$, which maps data points in a specific class to another fixed one. This means that the labels of selected instances in class $i$ are flipped to $r(i)$. To ensure distinctiveness, the mapping function $r$ adheres to the condition $r(i) \neq r(j)\; \forall i, j = \{1, ..., N\}$, and $i \neq j$. This noise models the situations where inputs from multiple annotators are often aggregated; the resulting differences in annotations can serve as a model of systematic noise derived from human disagreements.

\noindent (3) \textbf{\underline{Concentrated noise}}: we select data points that are densely clustered and change their labels to target labels.
We simulate scenarios where the datasets are poisoned by adversaries to evaluate the sanitization ability of methods against data poisoning attacks.

\paragraph{Settings and hyperparameters.} For each dataset, we construct groups of various sizes of noisy samples by corrupting the label of $p\%$ of the original training data. We construct the auxiliary dataset $\mathcal{D}_{\textnormal{aux}}$ by randomly selecting $m$ samples from the validation set. We fine-tune BERT on the noisy dataset $\mathcal{D}$ and select the best checkpoint measure on the validation set. We select the top $t\%$ ranked samples in $\mathcal{D}^{\uparrow}_{:p}$ and use detection accuracy for evaluation. After rectifying/removing ranked samples (potentially noisy samples), we re-train the model and report the test accuracy and noise reduction rate. Models were trained with AdamW~\cite{adamw} with learning rate $5e-5$, momentum $\beta = (0.9, 0.999)$, cross entropy loss, batch-size of $16$ with $15$ epochs.  For regularization, we use Dropout~\cite{dropout} of $0.2$. We choose $p=\{5\%, 10\%, 20\%\}$, $m=1000$, $k=\{1, 2, 5, 10, 20, 50, 100, 200\}$, $t=\{10\%, 20\%, ..., 100\%\}$, and $\tau=0.8$. We compute the IF score for BERT with the last layer gradient as suggested in previous works~\cite{pezeshkpour2021empirical, hanawa2020evaluation} and use LiSSA~\cite{agarwal2017second} to approximate the Hessian. For TracIn method, we calculate the influence score from the first checkpoint to the best checkpoint. We report results averaged over $4$ random seeds. Two Nvidia A40s, each $45$GB of GPU RAM, were used to run experiments.

\section{Result and Analysis}\label{sec:result}
\subsection{Main Result}
% \subsection{Noise Detection Performance} Fig.~\ref{fig2} shows the detection accuracy of methods with three types of noise with different percentages. As a result, when $t$ increases, the performance of gradient-based methods \textit{drastically decreases}. This pattern is observed in all three types of noise across different percentages and in both Snippets (Fig.~\ref{fig2}a-c) and IMDB (Fig.~\ref{fig2}d-f). This result shows that \textit{the gradient-based methods are unstable and inconsistent}. Confidence-based methods are precise with random noise and systematic ambiguity noise, yet struggle with concentrated noise. Sim-Cos and Sim-Dot achieved high detection accuracy and a slight decrease when $t$ increases with different noise. 
% This confirmed that the similarity-based methods are effective and more robust to ambiguity and concentrated noises than gradient-based and confidence-based methods.
\textbf{Similarity-based methods demonstrate superior robustness and effectiveness.} Fig.~\ref{fig2} shows the detection accuracy of different methods under three types of noise at varying noise levels. As $t$ increases, the performance of gradient-based methods \textit{drops significantly}, a pattern consistently observed across all noise types and percentages in both Snippets (Fig.~\ref{fig2}a–c) and IMDB (Fig.~\ref{fig2}d–f). These results indicate that \textit{gradient-based methods are unstable and inconsistent}. In contrast, confidence-based methods perform well under random and systematic ambiguity noise but struggle with concentrated noise. Both Sim-Cos and Sim-Dot achieve high detection accuracy with only a slight degradation as $t$ increases, demonstrating that \textit{similarity-based methods are more effective and robust to ambiguity and concentrated noise than gradient-based and confidence-based approaches}.

\noindent\textbf{Empirical validation for the claim in Section~\ref{theoretical_analysis}.} We observe that Sim-Dot and Sim-Cos consistently have lower detection accuracy on IMDB than that of Snippets (Fig.~\ref{fig2}). We theoretically proved that for classification datasets with $N$ classes, the similarity of within-class data points is approximately $N-1$ times larger than that of other class data points (c.f. Eqn.~\ref{eq22}). For IMDB, where $N=2$, this fraction becomes small. This explains why Sim-Dot and Sim-Cos achieve higher performance on Snippets than on IMDB. 

\noindent \textbf{Why Sim-Cos consistently outperforms Sim-Dot?} 
% \begin{figure}[!ht] 
% \centering 
% \includegraphics[width=0.45\textwidth]{images/l2_norm.png}
% \caption{.}
% \label{fig1}
% \end{figure}
We explain from the standpoint of \textit{feature normalization}. By definition, cosine similarity can be seen as the normalized dot product. We empirically observed that the noisy features' norm is smaller than that of benign samples. Therefore, when dividing the feature of data points by their norm, the similarity between noisy and normal data points tends to be larger, leading to a more distinct distribution of similarities. 
% \paragraph{Improving datasets' quality.} Tab.\ref{tab:1} shows a significant improvement in the test accuracy when removing/rectifying concentrated noise and systematic ambiguity noise. Nevertheless, a counterintuitive observation regarding random noise shows that removing/rectifying noise reduces the generalization of models, even when detection accuracy is high. We posit that deep models are robust to massive random noise~\cite{rolnick2017deep}, then as the training process continues, the model also memorizes the noise, approaches the optimum, and the gradient of noise becomes smaller. The effect of noise, therefore, also decreases as the model converges. When removing noise, the model degrades the feature representation of noise samples and loses the generalization to unseen samples.

\noindent \textbf{Models’ performance on noise-removed or noise-rectified datasets.} Tab.\ref{tab:1} and Tab.~\ref{tab:2} show the test accuracy of BERT trained on Snippets and IMDB after removing or rectifying potential noisy samples. We observe that for more harmful types of noise---such as ambiguity or concentrated noise---removing or rectifying these samples improves model generalization. In contrast, a counterintuitive trend emerges for random noise: removing or rectifying such noise tends to reduce generalization, even when the noise detection accuracy is relatively high. 
This phenomenon further highlights the notion that random noise can act as a useful regularizer for enhancing models' generalization, indicating that deep nets exhibit strong robustness even under random noise~\cite{rolnick2017deep}.
%We posit that deep nets are robust to massive random noise~\cite{rolnick2017deep}, then as the training process, the model also memorizes the noise, approaches the optimum, and the gradient of noise becomes smaller. The effect of noise, therefore, also decreases as the model converges. When removing noise, the model degrades the feature representation of noise samples and loses the generalization to unseen samples.

% \section{Ablation Study}
% \label{sec:appendixC} 
\subsection{Ablation Study}
\begin{table*}[!ht]
\centering
\small
\caption{Detection accuracy of Sim-Cos and Sim-Dot across noise types at different levels using three LLMs on Snippets and IMDB.}
\label{tab:improvement}
\resizebox{\linewidth}{!}{%
\begin{tabular}{@{}lcccccccccccccccccccccccccccc@{}}
\toprule
\multirow{3}{*}{\textbf{Model}} 
& \multicolumn{6}{c}{\textbf{Random noise}} 
& \multicolumn{6}{c}{\textbf{Ambiguity noise}} 
& \multicolumn{6}{c}{\textbf{Concentrated noise}} 
\\ \cmidrule(lr){2-7} \cmidrule(lr){8-13} \cmidrule(l){14-19}

& \multicolumn{2}{c}{\textbf{$5\%$ Noise}} 
& \multicolumn{2}{c}{\textbf{$10\%$ Noise}} 
& \multicolumn{2}{c}{\textbf{$20\%$ Noise}} 
& \multicolumn{2}{c}{\textbf{$5\%$ Noise}} 
& \multicolumn{2}{c}{\textbf{$10\%$ Noise}} 
& \multicolumn{2}{c}{\textbf{$20\%$ Noise}} 
& \multicolumn{2}{c}{\textbf{$5\%$ Noise}} 
& \multicolumn{2}{c}{\textbf{$10\%$ Noise}} 
& \multicolumn{2}{c}{\textbf{$20\%$ Noise}} 
\\ \cmidrule(lr){2-3} \cmidrule(lr){4-5} \cmidrule(lr){6-7}
\cmidrule(lr){8-9} \cmidrule(lr){10-11} \cmidrule(lr){12-13}
\cmidrule(lr){14-15} \cmidrule(lr){16-17} \cmidrule(l){18-19}

& Cos & Dot 
& Cos & Dot 
& Cos & Dot 
& Cos & Dot 
& Cos & Dot 
& Cos & Dot 
&Cos & Dot 
& Cos & Dot 
& Cos & Dot \\ 
\midrule

& \multicolumn{18}{c}{\textit{Snippets}} \\
BERT (trained)     & $85.3$ & $83.1$ & $89.2$ & $88.2$ & $93.1$ & $92.9$ & $86.7$ & $84.5$ & $90.5$ & $90.1$ & $92.9$ & $91.0$ & $47.6$ & $45.3$ & $64.0$ & $58.0$ & $79.5$ & $75.0$ \\
Llama-3.2-1B     & $45.0$ & $36.3$ & $57.2$ & $47.8$ & $66.8$ & $60.4$ & $42.7$ & $40.2$ & $61.0$ & $57.9$ & $72.3$ & $61.6$ & $38.2$ & $34.2$ & $58.0$ & $47.8$ & $72.8$ & $62.6$ \\
Llama-3.2-3B     & $47.2$ & $41.2$ & $56.5$ & $52.1$ & $66.8$ & $63.9$ & $40.7$ & $40.0$ & $60.0$ & $58.9$ & $71.9$ & $67.2$ & $37.0$ & $35.2$ & $56.2$ & $54.5$ & $72.6$ & $67.5$ \\
Llama-3.1-8B     & $51.2$ & $50.7$ & $61.9$ & $60.3$ & $71.4$ & $68.0$ & $43.2$ & $41.2$ & $65.0$ & $63.8$ & $75.4$ & $69.4$ & $41.7$ & $40.5$ & $63.1$ & $57.7$ & $79.8$ & $73.2$ \\
% Llama-3.1-70B    & $53.9$ & $50.7$ & --- & --- & --- & --- & --- & --- & --- & --- & --- & --- & --- & --- & --- & --- & --- & --- \\
\midrule

& \multicolumn{18}{c}{\textit{IMDB}} \\
BERT (trained)     & $65.0$ & $59.5$ & $73.2$ & $68.0$ & $78.8$ & $76.9$ & $67.8$ & $64.2$ & $73.0$ & $70.1$ & $78.8$ & $78.5$ & $55.6$ & $45.0$ & $63.2$ & $58.2$ & $60.0$ & $52.0$ \\
Llama-3.2-1B     & $11.7$ & $8.3$ & $21.1$ & $16.5$ & $35.0$ & $31.2$ & $10.8$ & $8.0$ & $20.5$ & $17.3$ & $33.8$ & $31.1$ & $12.2$ & $9.4$ & $22.3$ & $17.5$ & $35.7$ & $29.8$ \\
Llama-3.2-3B     & $21.4$ & $12.2$ & $27.9$ & $21.1$ & $39.3$ & $34.2$ & $21.9$ & $11.5$ & $25.8$ & $21.0$ & $38.1$ & $33.6$ & $20.9$ & $12.3$ & $29.7$ & $21.5$ & $40.2$ & $33.6$ \\
Llama-3.1-8B     & $24.2$ & $13.4$ & $29.5$ & $18.1$ & $41.9$ & $30.1$ & $23.5$ & $12.2$ & $29.0$ & $18.1$ & $41.2$ & $29.9$ & $22.9$ & $11.7$ & $31.7$ & $18.6$ & $42.6$ & $31.1$ \\
% Llama-3.1-70B    & --- & --- & --- & --- & --- & --- & --- & --- & --- & --- & --- & --- & --- & --- & --- & --- & --- & --- \\
\bottomrule
\end{tabular}
}
\end{table*}
\noindent\textbf{Effects of the size of auxiliary set.} 
\begin{figure}[!ht] 
\centering 
\includegraphics[width=0.23\textwidth]{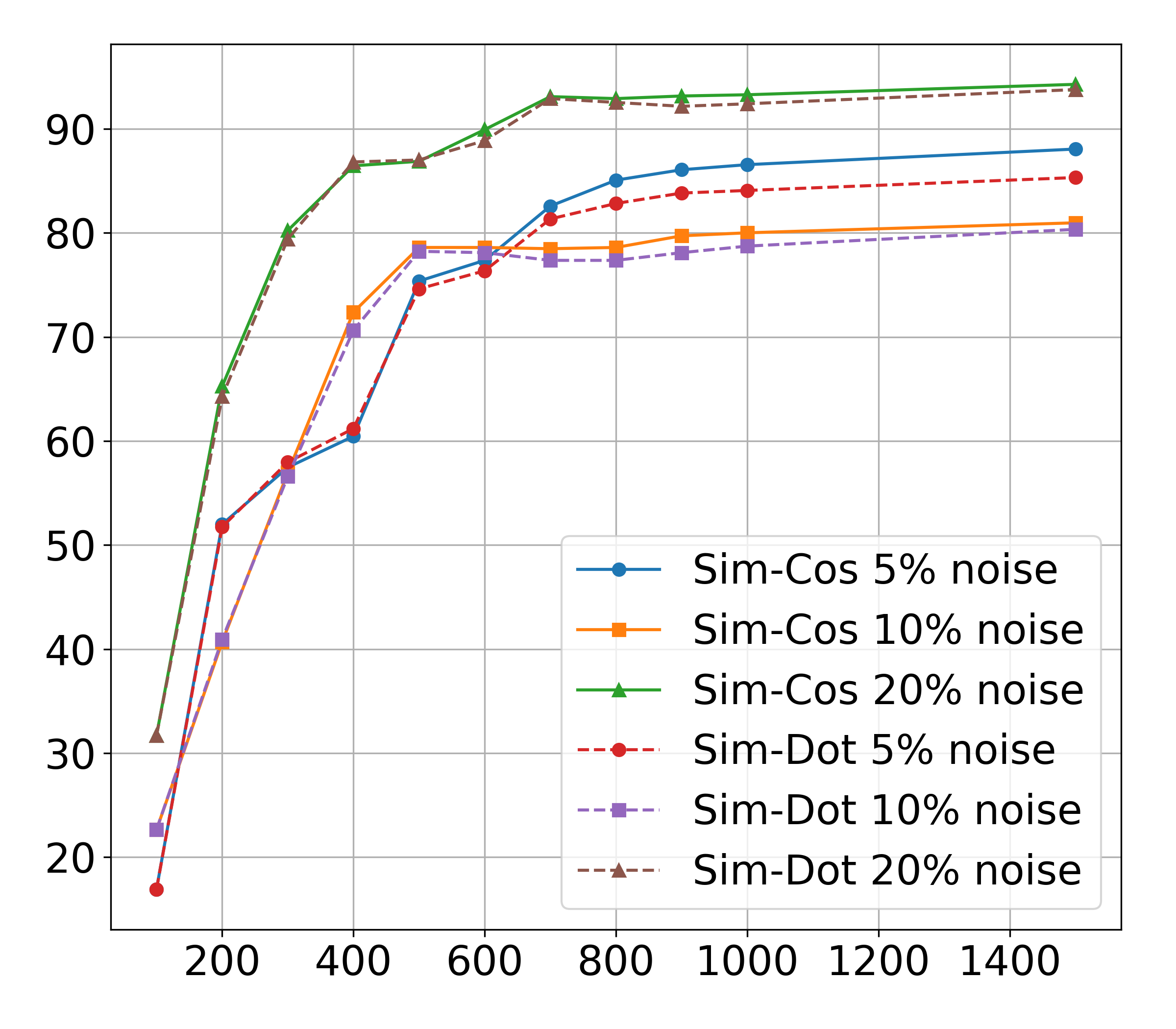}
\includegraphics[width=0.23\textwidth]{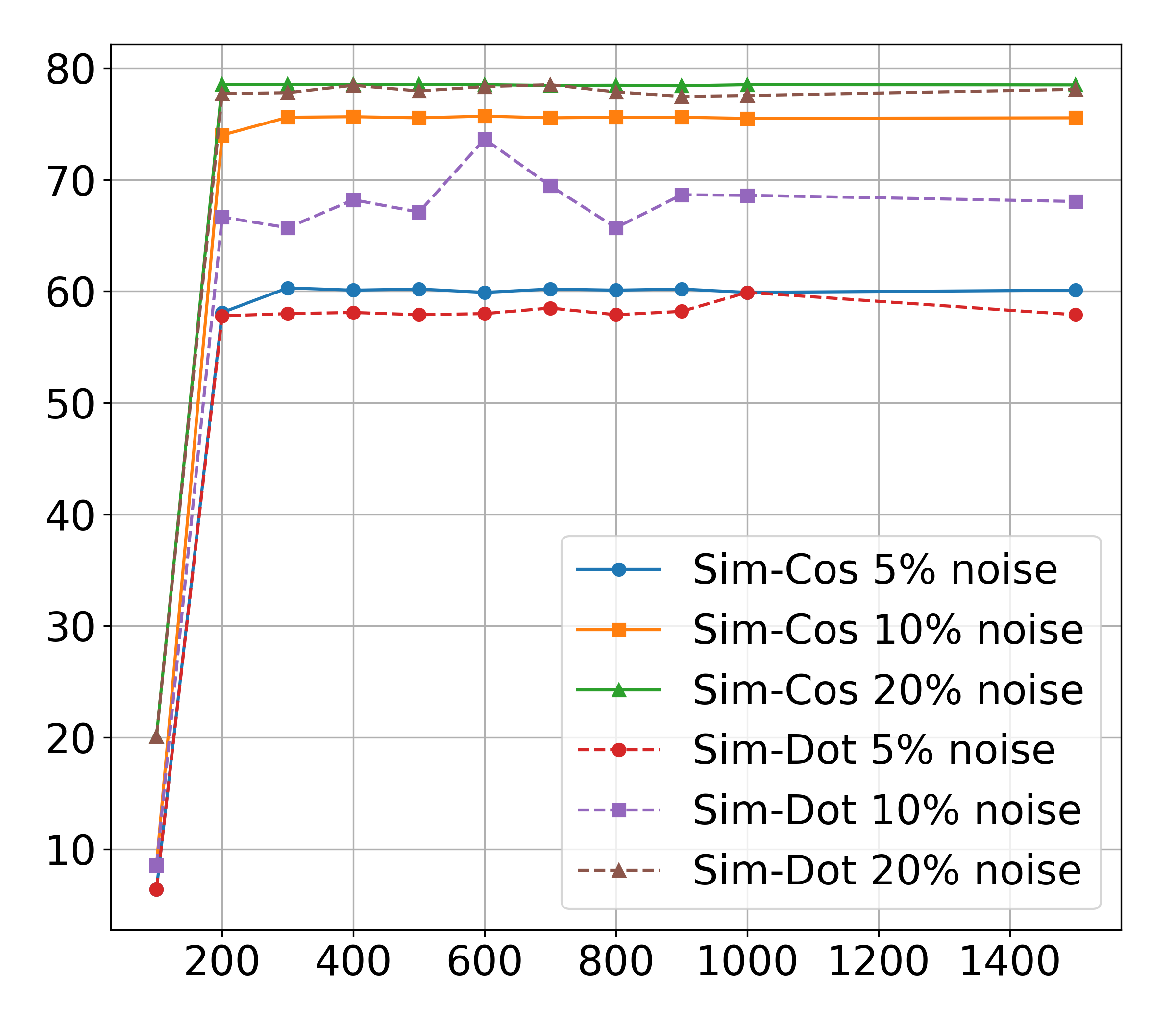}

\caption{Detection accuracy of models on Snippets (left) and IMDB (right) across sizes of $\mathcal{D}_{\textnormal{aux}}$.}
\label{fig3}
\end{figure}
Fig.~\ref{fig3} shows the change in the detection accuracy as the size of $\mathcal{D}_{\textnormal{aux}}$ increases from $100$ to $1500$, fixing $k=100$. We observed that: the detection accuracy of the methods increases as the size of $\mathcal{D}_{\textnormal{aux}}$ increases for both Snippets and IMDB, Sim-Cos and Sim-Dot, and for all levels of noise. Sim-Cos outperforms Sim-Dot for all noise levels.

\noindent\textbf{Effects of $k$ and $\tau$.} 
\begin{figure}[!ht] 
\centering 
\includegraphics[width=0.23\textwidth]{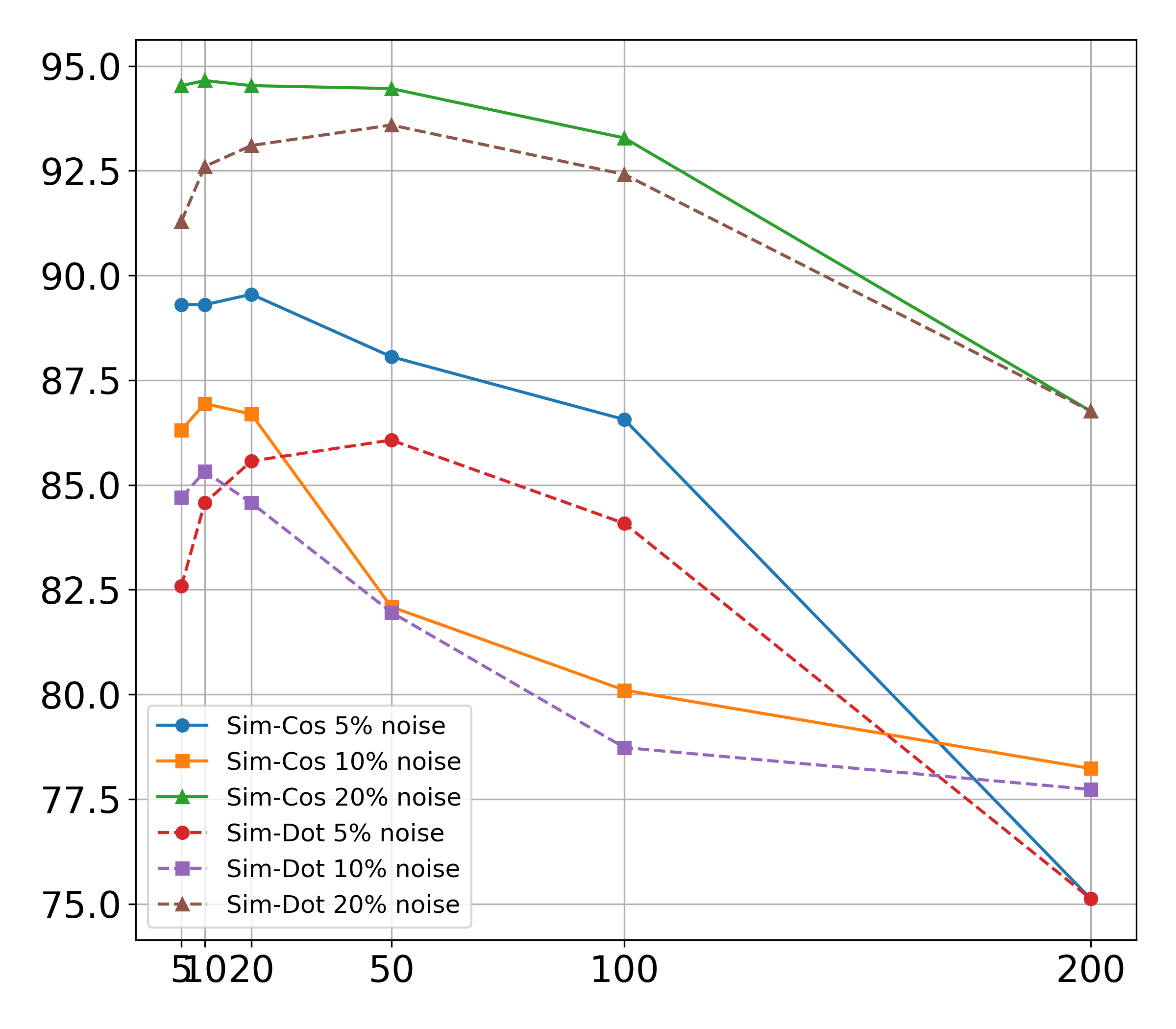}
\includegraphics[width=0.23\textwidth]{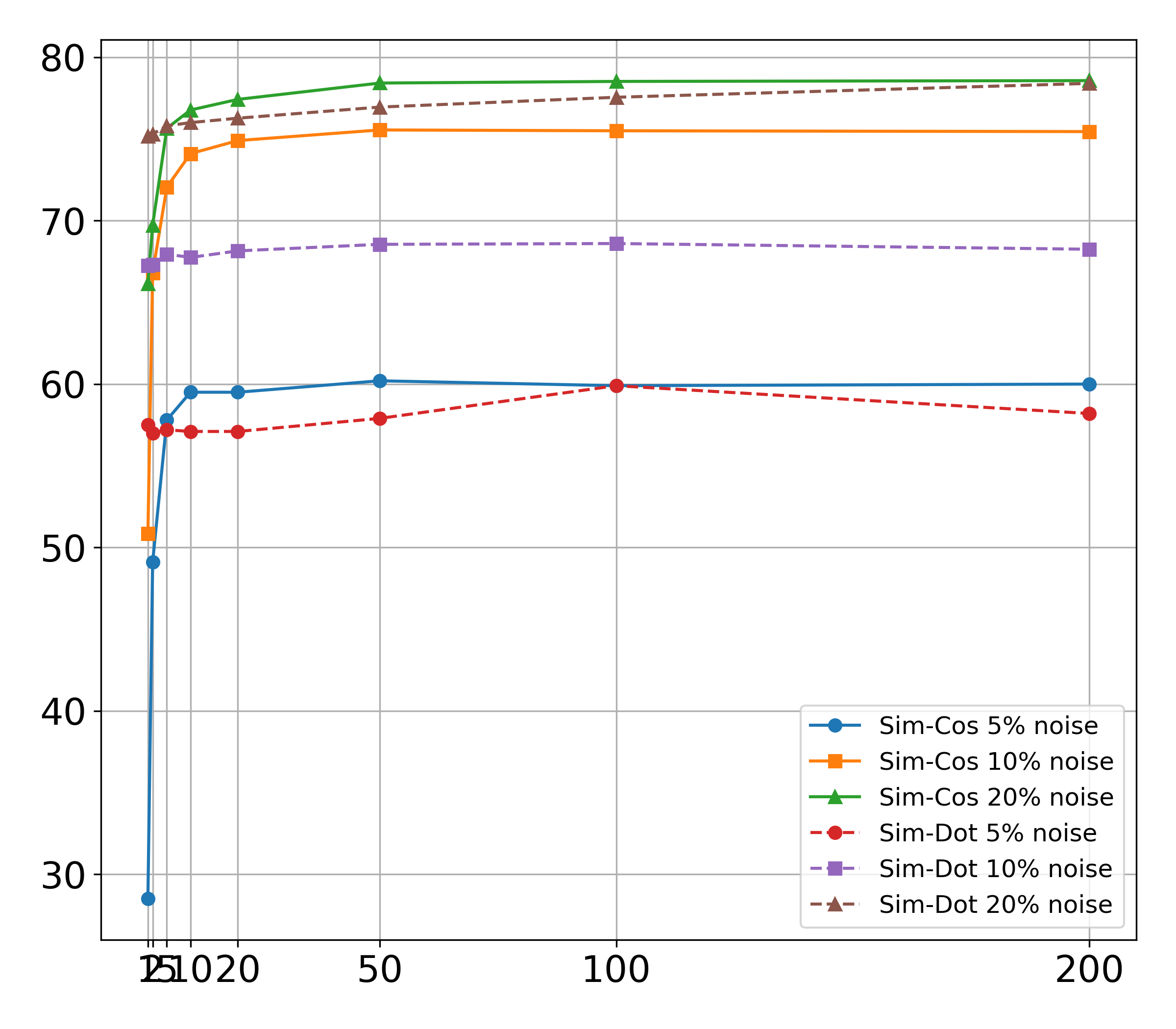}

\caption{Detection accuracy of similarity-based methods on Snippets (left) and IMDB (right) for $k \in \{5,10,20,50,100,200\}$.}
\label{fig4}
\end{figure}
Fig.~\ref{fig4} shows that Sim-Cos consistently outperforms Sim-Dot as $k$ increases for all noise levels. The impact of $k$ on detection accuracy with Snippets is generally small; with performance beginning to decline beyond $k = 20$. This suggests that using a larger $k$ may degrade the detection accuracy. For IMDB, the accuracy increases as $k$ increases, but quickly reaches a saturation point at $k$ around $20$. We analyze the effect of $\tau$ on the error reduction rate. Fig.~\ref{fig5} shows that for both Snippets and IMDB, the change of $\tau$ has minimal impact on the error reduction rate. Generally, the reduction rate is higher when the noise level is higher.

\begin{figure}[!ht] 
\centering 
\includegraphics[width=0.23\textwidth]{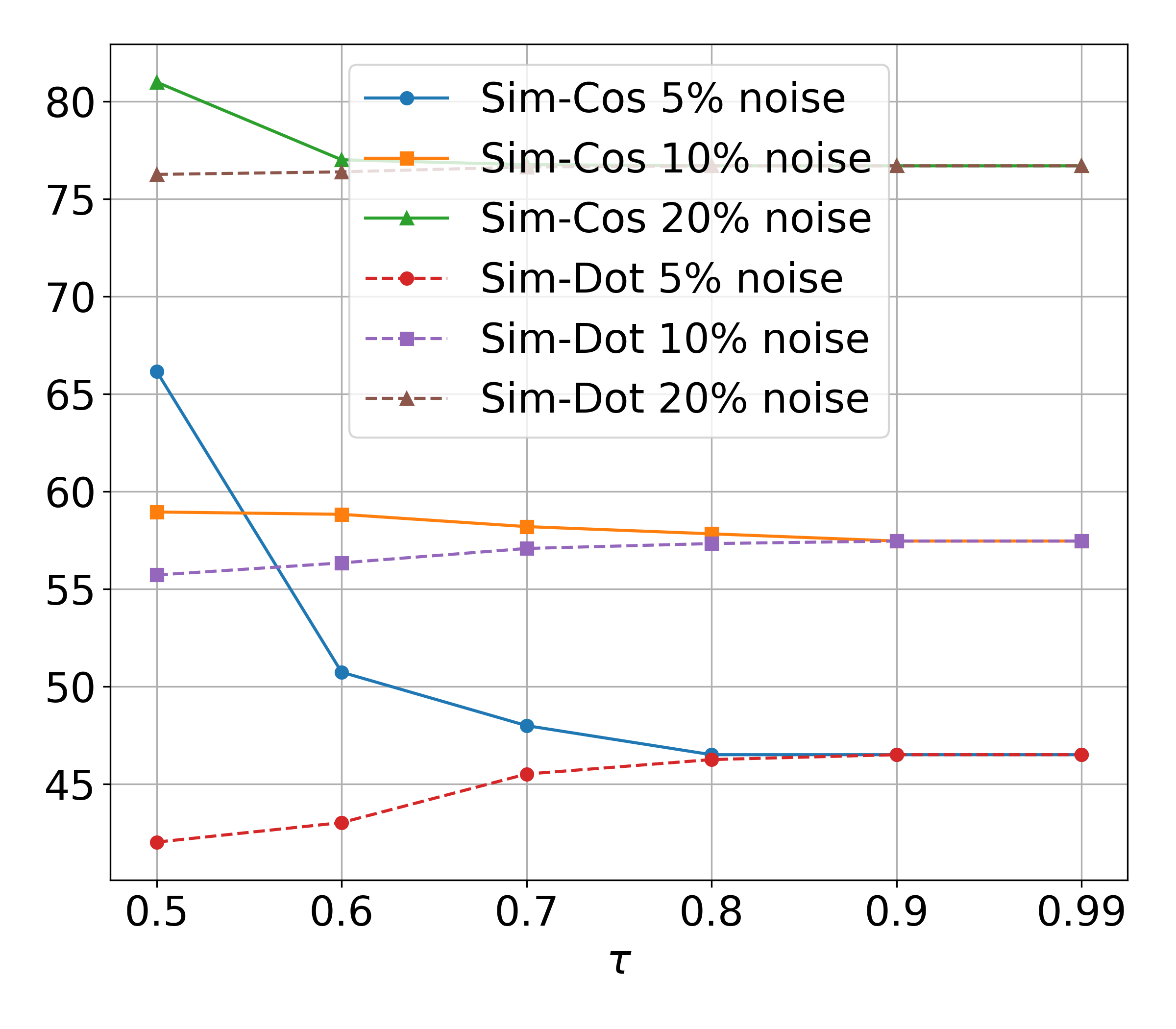}
\includegraphics[width=0.23\textwidth]{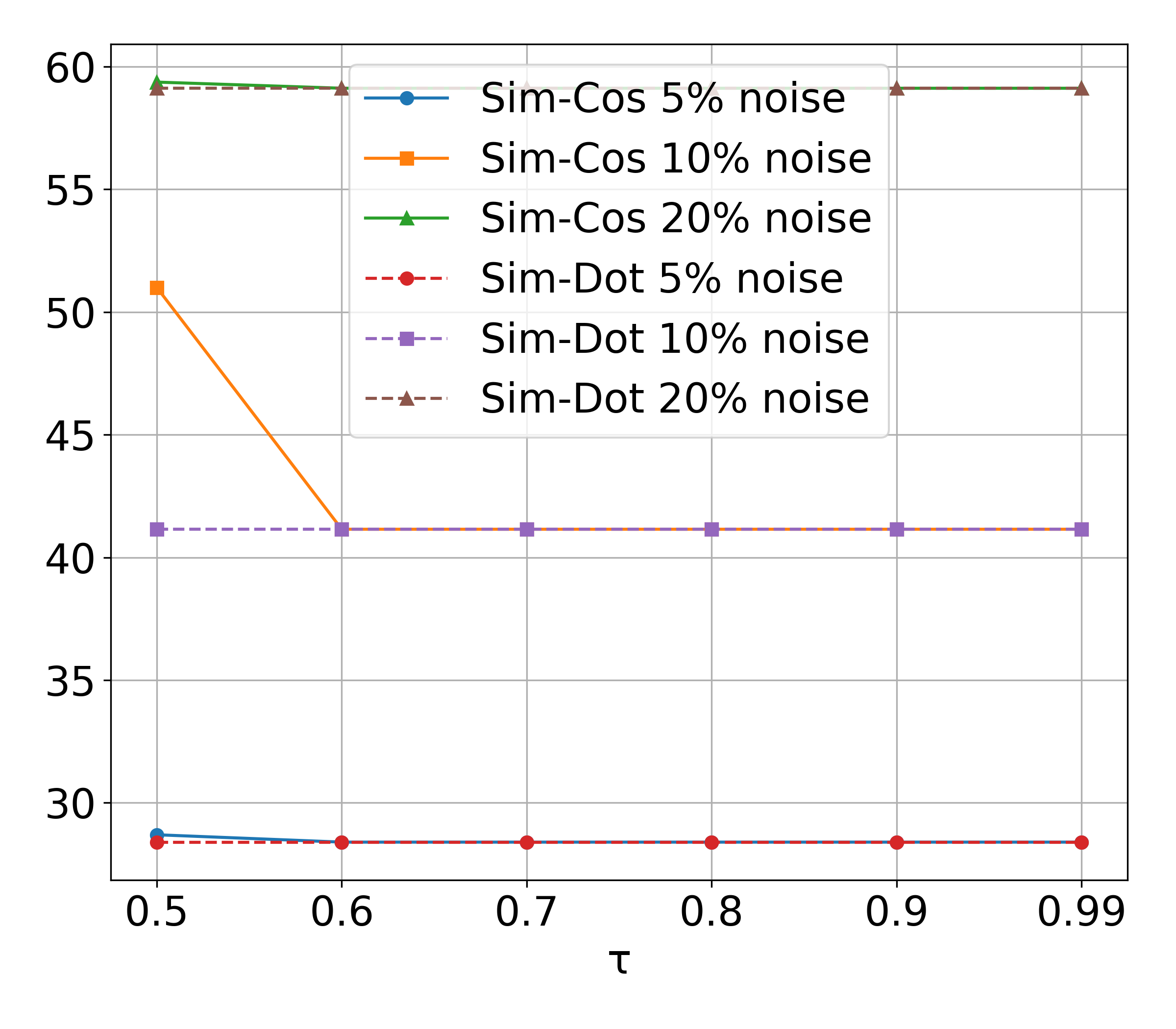}
\caption{Noise reduction rate across threshold $\tau \in \{0.5,0.6,0.7,0.8, 0.9, 0.99\}$.}
\label{fig5}
\end{figure}

\noindent \textbf{Generalization to LLMs.}\label{zero_shot} The size of the model and the diversity of its pretraining data endow LLM with strong capabilities on downstream tasks, even without explicit task-specific fine-tuning~\cite{NEURIPS2020_1457c0d6}. 
% Our methods are built based on the observation that the mislabeled data points
% are often more similar to true class data points than
% those from other classes in the \textit{trained} model's penultimate representation space. 
One might ask: \textit{``Given a general-purpose LLM, which has not been trained on noisy datasets, how effectively can the proposed methods perform?''} We aim to evaluate the effectiveness of similarity-based methods with three open-weight LLMs, including Llama-3.2-1B, Llama-3.2-3B, and Llama-3.1-8B. Tab.~\ref{tab:improvement} shows the detection accuracy across different noise types and levels.
Overall, Sim-Cos consistently outperforms Sim-Dot across all settings. However, both Sim-Cos and Sim-Dot achieve lower absolute accuracy with LLMs compared to the fine-tuned BERT model. This can be intuitively explained by the fact that, in a well-trained BERT, features from different classes tend to be pushed into different orthogonal subspaces (c.f. Section~\ref{theoretical_analysis} and~\citet{csordasneural}). In contrast, LLMs are not task-specific trained, causing features of different classes to be more entangled, which reduces the effectiveness of similarity-based methods.

\section{Conclusion}
We introduce similarity-based algorithms with theoretical justification for detecting and rectifying noise in datasets. Experiment results demonstrated the superior performance of our approach to improve the dataset quality and model generalization.
\newpage

\section*{Limitations}
We discuss the limitations of similarity-based methods:
(1) The optimal detection accuracy of Sim-Cos and Sim-Dot, unfortunately based on empirical validation and depends on the choice of $k$ and $\mathcal{D}_{\textnormal{aux}}$, and also with different datasets and model architectures.
(2) The generalization of models under the removal or rectification of noise remains uncertain, due to the limited exploration of datasets.
\section*{Ethics Statement}
We consider only the public datasets and create artificial noises for evaluation. We do not pose any concern about the quality of the original datasets. AI tools were used for grammar checking, and technical content was written by the authors.
% \subsection{Appendices}

% \subsection{Extra space for ethical considerations and limitations}

% Please note that extra space is allowed after the last page for an ethics/broader impact statement and a discussion of limitations. At submission time, if you need extra space for these sections, it should be placed after the conclusion so that it is possible to rapidly check that the rest of the paper still fits in 8 pages maximum. Ethical considerations sections, limitations, acknowledgments, and references do not count against these limits.

% \section{Providing References}

% \subsection{Bibliographical References} 

% Bibliographical references should be listed in alphabetical order at the end of the paper. The title of the section, ``Bibliographical References'', should be a Level 1 Heading. The first line of each bibliographical reference should be justified to the left of the column, and the rest of the entry should be indented by 0.35 cm.

% The examples provided in Section~\ref{sec:reference} (some of which are fictitious references) illustrate the basic format required for papers in conference proceedings, books, journal articles, PhD theses, and books chapters.

% \subsection{Language Resource References}

% Language resource references should be listed in alphabetical order at the end of the paper.

% \nocite{Eco:1990,Martin-90,Chercheur-94,CastorPollux-92,bs-2570-manual,bs-2570-techreport,chomsky-73,hoel-71-portion,hoel-71-whole,singer-whole,Jespersen:1922,Superman-Batman-Catwoman-Spiderman-00}
% \section{Bibliographical References}\label{sec:reference}

\bibliographystyle{lrec2026-natbib}
\bibliography{lrec2026-example}

% \section{Language Resource References}
% \label{lr:ref}
\bibliographystylelanguageresource{lrec2026-natbib}
\bibliographylanguageresource{languageresource}

\end{document}